\documentclass[pdflatex,sn-mathphys-num]{sn-jnl}

\usepackage{amsmath, amssymb}
\usepackage{graphicx}
\usepackage{booktabs}
\usepackage{siunitx}
\usepackage{url}
\usepackage{float}
\usepackage{hyperref}
\usepackage{algorithm}      
\usepackage{algpseudocode}
\usepackage{tikz}
\usetikzlibrary{arrows.meta,positioning,shapes.geometric}
\usepackage{booktabs}
\usepackage{multirow}
\usepackage{tabularx}
\usepackage{graphicx}
\usepackage{makecell}
\usepackage{lmodern}
\usepackage{float}
\theoremstyle{thmstyleone}%
\theoremstyle{thmstyletwo}%

\theoremstyle{thmstylethree}%

\begin{document}

\title[Article Title]{Beyond the Graph: An Adaptive Meta-Learner Fuses Explainability, Weather, and Dynamics for Robust Bus ETA Prediction}


\author*[1]{\fnm{Pratham} \sur{Payra}}\email{prathamfbm@gmail.com}

\author[2]{\fnm{Jagadish} \sur{ }}\email{jagadish@isibang.ac.in}
\equalcont{These authors contributed equally to this work.}

\affil*[1]{\orgdiv{SQC and OR division}, \orgname{Indian Statistical Unit}, \orgaddress{\street{BT Road}, \city{Kolkata}, \postcode{700108}, \state{West Bengal}, \country{India}}}

\affil[2]{\orgdiv{SQC and OR unit}, \orgname{Indian Statistical Unit}, \orgaddress{\street{8th Mile, Mysore Road}, \city{Bangalore}, \postcode{560059}, \state{Karnataka}, \country{India}}}


\abstract{Accurate bus Estimated Time of Arrival (ETA) prediction is vital for urban mobility, passenger satisfaction, and transit efficiency, yet existing models falter against nonlinear spatiotemporal dynamics, data sparsity, and factors like weather. This paper proposes HYB($n_m$), an adaptive hybrid ensemble framework that dynamically fuses five complementary models—a historical baseline (MST-AV), periodical temporal pattern analysis (GDRN-DFT), Koopman Neural Operators for nonlinear dynamics (KOOP-NET), weather-integrated feature-engineered neural networks (FENN), and real-time graph convolutional networks (MGCN)—via a meta-learner attuned to real-time context. Evaluated on GPS and weather data from three Kolkata bus routes (more than 4,000 trips), it leverages individual strengths (e.g., MST-AV's low-latency explainability, FENN's weather resilience, MGCN's network dynamics) to yield HYB(2)'s superior robustness, state-of-the-art accuracy rivaling top graph neural networks, and balanced trade-offs in stability and efficiency across horizons and conditions. The extensible HYB(k) architecture equips agencies with flexible tools, from economical single models to tailored high-fidelity hybrids, advancing predictive, equitable urban transport.}

\keywords{Bus arrival time prediction, Hybrid ensemble learning, Spatio-temporal forecasting, Intelligent transportation systems, Graph neural networks, Koopman operator}



\maketitle

\section{Introduction}\label{sec1}

Precise bus arrival time prediction forms the bedrock of reliable public transportation, which is, in turn, fundamental for sustainable urban mobility, economic vitality, and enhanced passenger experience \cite{Setiawan2024Integration}. The field has undergone a significant transformation, evolving from rudimentary static schedules to sophisticated data-driven models, largely propelled by the integration of GPS data. This progression has firmly established bus arrival time prediction as a pivotal research area within intelligent transportation systems \cite{Yuan2021ASO}. Nevertheless, as highlighted by a comprehensive review, existing models continue to contend with persistent limitations, including challenges posed by data sparsity and an insufficient capacity to fully capture complex urban traffic dynamics \cite{Kumar2025Bus}. Such shortcomings frequently lead to unreliable estimates, which can adversely affect both passenger satisfaction and overall ridership.

Consequently, this study aims to develop a novel hybrid framework specifically tailored for Kolkata's intricate bus system, integrating diverse data sources with advanced modeling techniques. The core objectives involve development of a robust spatiotemporal network model and a dynamic hybrid ensemble, leveraging the power of graph-based convolutional networks \cite{DBLP:journals/corr/abs-2007-02842, DBLP:journals/corr/abs-1909-11197} and neural operators \cite{DBLP:journals/corr/abs-2111-13587} to significantly enhance predictive accuracy. It is hypothesized that this integrated approach will lead to a substantial reduction in prediction errors (H1), that a superior capability to model nonlinear traffic patterns compared to traditional baselines will be demonstrated (H2), and that overall superior predictive performance across varied operational scenarios will be achieved by the proposed hybrid ensemble (H3). The current study offers a lot of potential, since it provides a computationally effective solution for transit operators while also significantly furthering the conversation on smart urban mobility.

\section{Related Work}\label{related_work}

The pursuit of accurate bus travel time prediction is a cornerstone of intelligent transportation systems, driven by its critical role in enhancing service reliability and passenger satisfaction. The field has witnessed a clear evolutionary trajectory, where each new methodological wave seeks to address the limitations of its predecessors.\\

The initial framework for ETA prediction was established through statistical models. Kumar et al. \cite{kumar2019seasonal} endeavoured to model recurring traffic patterns through seasonal spatio-temporal modelling, effectively capturing long-term trends. However, this approach faced challenges in handling complex, non-linear urban traffic dynamics. To model temporal patterns in automated contexts, Antypas et al. \cite{antypas2024time} applied time-series analysis with gradient boosting, yet it struggled to capture spatial interactions. To address the issue of data uncertainty, Chen et al. \cite{chen2023probabilistic} introduced a probabilistic Bayesian framework, providing full predictive distributions. While this improved reliability quantification, the problem of computational intensity persisted. For handling noisy data, Hsu et al. \cite{hsu2023hypereta} utilized hypercube clustering to improve data segmentation, though its clusters faced generalisability issues.\\

Concurrently, machine learning methods gained prominence to model more complex dynamics. Kam et al. \cite{kam2024predicting} leveraged boosting models to deliver strong baseline performance, yet a significant gap remained in their ability to model spatial dependencies within the transit network. Arifi et al. \cite{arifi2024study} tackled incomplete trajectories by using route recovery with classical ML, but its accuracy was inherently tied to the quality of the reconstructed paths. Similarly, Guo et al. \cite{guo2021integration} and Noor et al. \cite{noor2020predict} developed feature-engineered and classical ML models, respectively, prioritising interpretability and accessibility. Hassannayebi et al. \cite{hassannayebi2023data} engineered practical ANN pipelines, yet these models were fundamentally limited in graph modelling. Phon-Amnuaisuk et al. \cite{phon2023non} devised a non-GPS solution using historical contexts, but it relied heavily on hand-crafted features. Suresh et al. \cite{suresh2023estimation} leveraged user-location tracking for real-time estimation, introducing privacy concerns. A broader perspective was provided by Shanthi et al. \cite{shanthi2022analysis} through a data-mining analysis, though it lacked deep model comparisons.\\

The field shifted significantly with deep learning to automatically learn complex patterns. To overcome the temporal modelling limitations of prior methods, Bhutani et al. \cite{bhutani2024seq2seq} employed Sequence-to-Sequence RNNs, successfully capturing intricate temporal dependencies. Despite this advancement, challenges with long-range dependencies remained. Barnes et al. \cite{barnes2020bustr} and Tran et al. \cite{tran2020deeptrans} integrated neural sequence models with traffic forecasts for strong performance, but this introduced complexity and a dependence on external data. Chu et al. \cite{chu2023deep} tackled feature interaction with a deep encoder cross-network, while Aditya et al. \cite{aditya2022eta} and Munkhbayar et al. \cite{munkhbayar2025deep} demonstrated the efficacy of modern deep architectures,however the issue of explicitly encoding the road network's structure was unaddressed. Sun et al. \cite{sun2020codriver} incorporated driver behavior via auxiliary learning, improving personalization but raising data privacy issues. Paliwal et al. \cite{paliwal2019each} focused on route-specific accuracy with generative modeling, at the cost of requiring separate models per route. To mitigate deployment challenges, Fu et al. \cite{fu2020compacteta} developed compact ETA for fast inference, though it increased engineering complexity.\\

In previous works neglect of spatial topology became the focus of graph-based research. Sharma et al. \cite{sharma2023estimating} and Porvatov et al. \cite{porvatov2021hybrid} utilised Graph Convolutional Networks and hybrid graph embeddings, respectively, to directly model the transportation network, thereby improving spatial correlation capture. Reich et al. \cite{reich2022attention} further enhanced this with attention mechanisms to capture long-range network dependencies. However, these graph-based solutions introduced new difficulties in the directions of scalability, computational efficiency, and inference costs, limiting their real-time application.\\

The development of hybrid frameworks has emerged from this ongoing cycle of problem-solving. That aim to synthesize the strengths of multiple paradigms. Chen et al. \cite{chen2022hseta} proposed the HSETA framework, which combines heterogeneous data sources to improve robustness, particularly with sparse data. Meanwhile, Wang et al. \cite{wang2018learning} architected a Wide-Deep recurrent hybrid to blend memorisation and generalisation capabilities. Barnes et al. \cite{barnes2020bustr} also demonstrated the power of hybrid ensembles for robustness. While these hybrid models often achieve state-of-the-art accuracy, they culminate in a trade-off, facing persistent problems related to system complexity, challenging training pipelines, and interpretability, presenting a significant hurdle for practical, large-scale deployment.

\subsection*{2.1. Summary of Literature and Proposed Solutions}
Prior approaches to bus travel time prediction, encompassing statistical, machine learning, deep learning, graph-based, and hybrid models, face significant limitations in addressing the complexities of urban transportation dynamics. These include inadequate handling of spatial discretisation, nonlinear temporal patterns, and weather-related factors, as well as challenges in modelling periodic spatiotemporal behaviors and sparse data. Additionally, the lack of real-time data assimilation, limited model explainability, and high computational demands hinder their practical application. To address the previous issues, the present work introduces novel methodologies which not only overcome present gaps but also enhance prediction accuracy and deployment feasibility. The proposed study increases ETA prediction by increasing spatial and temporal modeling, adding contextual data, and solving computational and interpretability difficulties. The table~\ref{tab:limitations} provides a comprehensive overview of these limitations and details how the current methodologies systematically address them.

\begin{table}[t]
\centering
\caption{Summary of Limitations in Prior Studies and Mitigation Strategies Employed by the Proposed Framework}

\label{tab:limitations}
\begin{tabularx}{\textwidth}{p{3.5cm} p{3.0cm} X}
\toprule
\textbf{Limitations / Gaps} &
\textbf{Methods Used} &
\textbf{Addressing the Limitation} \\
\midrule

Lack of spatial discretisation &
MST-AV with grid approximation &
Continuous GPS trajectories are discretised using $50\,\text{m}\times50\,\text{m}$ grids with median referencing, enabling graph-based modelling and noise reduction. \\

Absence of non-linear temporal modelling &
Koopman-based KNN &
Non-linear bus-speed dynamics are captured via linear evolution in a lifted latent space using the Koopman operator. \\

No incorporation of weather effects &
FENN with feature engineering &
Real-time weather variables from OpenWeatherMap are encoded to improve contextual forecasting. \\

Inadequate modeling of spatio-temporal periodicity &
GDRN-DFT &
Graph diffusion combined with DFT captures daily and weekly traffic cycles effectively. \\

Poor handling of missing data &
GDRN with Bi-LSTM &
Spatial diffusion and bidirectional temporal modeling enable effective imputation of missing speeds. \\

Lack of real-time message passing &
Masked GCN &
Observed speeds are propagated across the graph in real time to complete network-wide estimates. \\

Limited explainability &
Hybrid stacking framework &
A meta-learner assigns interpretable weights to individual models based on context. \\

High computational cost &
$HYB(n_m)$ downsizing &
Dynamic model selection activates only the top-$n_m$ contributors, reducing inference cost. \\

Unbounded predictions &
Koopman + FENN &
Logit and sigmoid transforms constrain predicted speeds within feasible limits. \\

\bottomrule
\end{tabularx}
\end{table}

\section{Research Methodology}\label{research_methodology}
\begin{figure}[h]
    \centering
    \includegraphics[width=.9\textwidth]{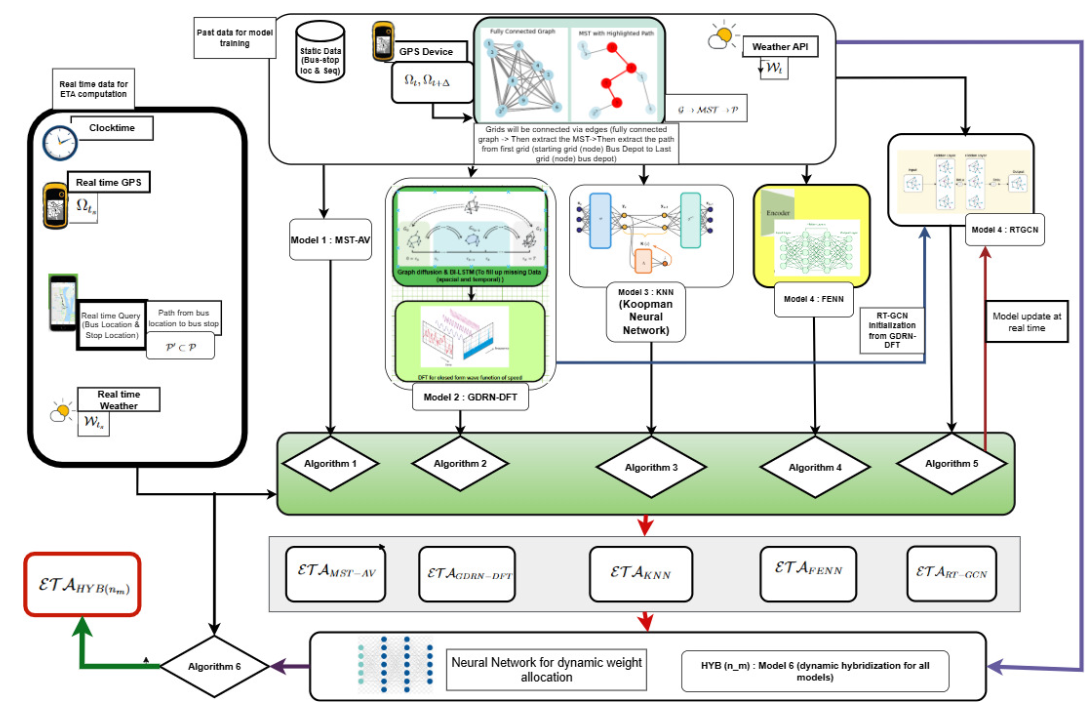}
    \caption{Overview of the proposed methodology}
    \label{fig:methodology_overview}
\end{figure}

The present work aims to develop a highly robust Bus Arrival Time Prediction (BATP) system through the systematic integration of high-frequency, granular spatiotemporal bus operational data with crucial contextual environmental factors. The proposed methodology, shown in Figure~\ref{fig:methodology_overview}, consists of three phases: 1) Data Acquisition and Preprocessing to establish foundational inputs, 2) Spatiotemporal Network Construction to map transport dynamics, and 3) Hybrid Prediction Framework for advanced inference. This proposed integrated design is specifically chosen to directly address the paramount challenge within BATP: accurately modelling the extremely complex and often non-linear interplay between a vehicle's dynamic physical path, predictable recurrent traffic patterns, and the unpredictable non-recurrent disruptions caused by stochastic exogenous factors such as varying meteorological conditions. The advanced hybrid framework is strategically posited to synergistically combine the complementary strengths of mechanistic, statistical, and deep learning paradigms, thereby achieving superior predictive robustness and accuracy across a wide spectrum of dynamic and diverse operational conditions.\\

\subsection{Data Acquisition and Preprocessing}
The present work used three primary categories of data sources to enable effective development of a robust BATP system. These categories were carefully selected for their direct alignment with the established, fundamental dimensions critically important for building resilient BATP models: specifically, the intricate network characteristics, essential temporal dynamics, and vital contextual factors. This careful selection strategy ensures the construction of a holistic data foundation for all subsequent analytical and predictive processes.\\
\\The data sources utilised include:\\\\
\textbf{Static Bus Stop Data:} Data includes geo-referenced records detailing the precise locations of all bus stops, which serve as key nodes and crucial prediction targets within the constructed transportation network.\\\\
\textbf{Historical Weather Data:} Data encompasses parameters such as atmospheric temperature, relative humidity, precipitation levels, wind speed, and cloud cover, recorded at hourly intervals. These environmental factors are widely documented as significant covariates influencing traffic flow and, consequently, bus arrival times.
Following data acquisition, the raw Global Positioning System data, frequently characterised by inherent noise and susceptibility to outliers, was subjected to a rigorous and meticulously designed preprocessing pipeline. A foundational step involved the precise calculation of instantaneous speed between successive geographic points utilising the Haversine formula, which accurately computes the great-circle distance. Furthermore, a crucial procedural stage, vital for enabling both efficient spatial analysis and robust graph-based modeling, encompassed the discretisation of the continuous coordinate space. This was achieved by segmenting the area into a regular grid, with each cell possessing approximate dimensions of \SI{50}{m} $\times$ \SI{50}{m}. This strategic approach not only significantly reduces computational complexity but also diligently preserves a sufficient degree of geographic fidelity. To effectively mitigate the pervasive impact of GPS noise and ensure data stability, the median coordinates of all GPS points situated within a given cell were subsequently computed. These median values served to define a stable, highly representative node for each respective cell within the discretised network.
\begin{figure}[h]
    \centering
    \includegraphics[width=.9\textwidth]{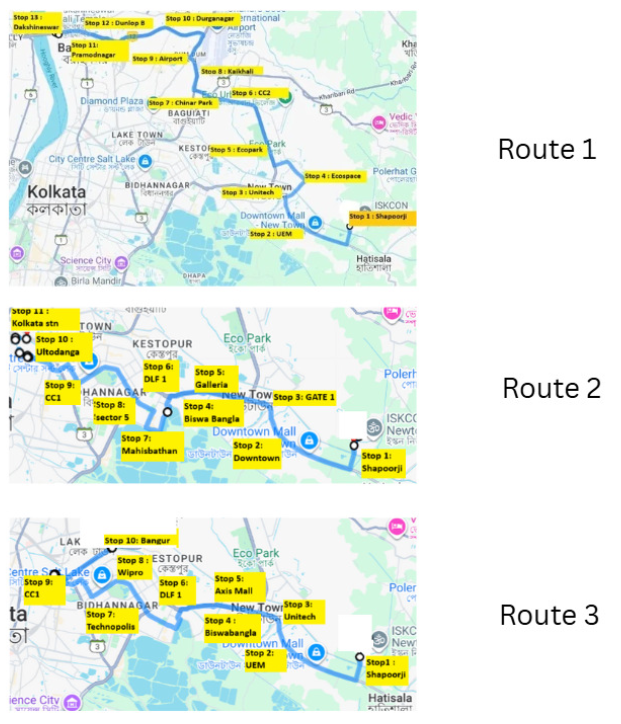}
    \caption{Selected routes for ETA computation}
    \label{fig:Routes selected for ETA computation}
\end{figure}\\


\begin{table}[h!]
\centering
\setlength{\tabcolsep}{1pt}
\caption{Specific Route GPS data and Statistics}
\label{tab:route_specific_stat}
\fontsize{4.75}{5}\selectfont
\begin{tabular}{@{}lccc@{}}
\toprule
 & \multicolumn{3}{c}{\textbf{Routes}} \\
\cmidrule(lr){2-4}
\textbf{Metric} 
& \textbf{Route 1} 
& \textbf{Route 2} 
& \textbf{Route 3} \\
& \textbf{(DN2/1: Shapoorji--Dakshineswar)} 
& \textbf{(KB22: Shapoorji--Kolkata Station)} 
& \textbf{(KB16: Shapoorji--Bangoor)} \\
\midrule
Total Sessions 
& 1,251 & 1,829 & 953 \\
Total GPS Points 
& 45,321 & 62,553 & 35,268 \\
Distance (km) 
& $26$ & $22$ & $19$ \\
Number of Stops
& $13$ & $11$ & $10$ \\
Avg.\ Speed (km/h) 
& $18.3 \pm 2.1$ & $15.7 \pm 1.8$ & $20.1 \pm 2.4$ \\
Median Speed (km/h) 
& 17.8 & 15.2 & 19.5 \\
\bottomrule
\end{tabular}
\end{table}

\begin{table}[h!]
\centering
\setlength{\tabcolsep}{40pt}
\caption{Table shows an even distribution across the study period, with slight weighting toward monsoon months (Aug–Oct) for higher collection frequency due to weather covariate focus. Total: 100\%.}
\label{tab:monthly_distribution}
\fontsize{6}{7}\selectfont
\begin{tabular}{@{}lcc@{}}
\toprule
\textbf{Month} 
& \textbf{\% of Total Sessions} 
& \textbf{\% of Total GPS Points} \\
\midrule
Aug 2024 & 16\% & 15\% \\
Sep 2024 & 17\% & 18\% \\
Oct 2024 & 15\% & 16\% \\
Nov 2024 & 13\% & 13\% \\
Dec 2024 & 12\% & 12\% \\
Jan 2025 & 11\% & 11\% \\
Feb 2025 & 9\%  & 9\%  \\
Mar 2025 & 7\%  & 6\%  \\
\bottomrule
\end{tabular}
\end{table}

\textbf{GPS Trajectory Data:} This category comprises granular records of bus movements, encompassing precise timestamps($\mathcal{T}_s$), geographic coordinates (latitude $\vartheta$, longitude $\varphi$), a unique session ID($\mathcal{S}$), and the associated route number($\mathcal{R}$).(Gps dataset contains: $\mathcal{D}_{gps} = \{\mathcal{S},\vartheta,\varphi,\mathcal{R},\mathcal{T}_s\}$) In the present work, real-time GPS data was collected over an extensive period, spanning from August 2024 to March 2025. Data collection focused on three specific public bus routes operating within Kolkata:(Shown in Figure~\ref{fig:Routes selected for ETA computation}) route 1, identified as DN-2/1 (traversing from Shapoorji to Dakshineswar); route 2, designated KB-22 (connecting Shapoorji to Kolkata Station); and route 3, labeled KB-16 (operating between Shapoorji and Bangoor). The geographical representation of these critical routes is visually presented in an accompanying figure. This comprehensive dataset ultimately forms the core spatiotemporal stream for our analysis.(Gps data specific information is given in Tables: ~\ref{tab:route_specific_stat} and ~\ref{tab:monthly_distribution})

\subsection{Spatiotemporal Network Construction}
\begin{figure}[htbp]
    \centering
    \includegraphics[width=1\textwidth]{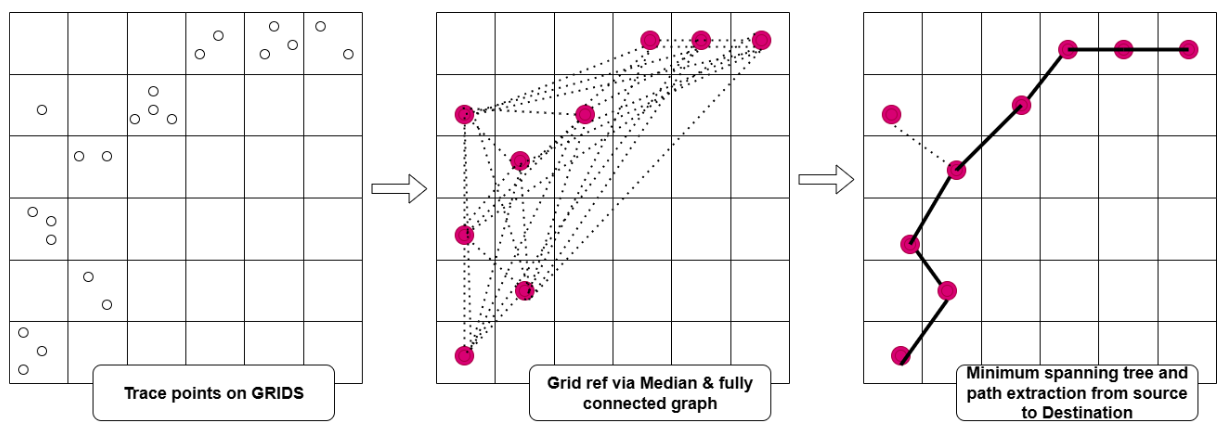}
    \caption{Graph Extraction Process (see Algorithm~\ref{alg:data_acq_preproc}.)}
    \label{fig:graph_construction}
\end{figure}

The discretised grid cells, previously generated from the GPS data, form the foundational vertex set $\mathcal{V}$ of a directed graph $\mathcal{G} = (\mathcal{V}, \mathcal{E})$, which meticulously represents the underlying transportation network. This sophisticated graph-based abstraction constitutes a fundamental conceptual pillar in network-based travel time estimation methodologies. Initially, a comprehensive complete graph was constructed, wherein the edge weights connecting any two nodes, $i$ and $j$, were precisely defined by the Haversine distance calculated between their respective median coordinates. To derive a plausible, yet parsimonious and sparse, representation of the actual road network topology from this dense point cloud, a Minimum Spanning Tree (MST) \cite{jain2023minimum} was subsequently extracted from this initially complete graph.(See figure~\ref{fig:graph_construction})The MST offers an elegantly connected graph that inherently minimises the total cumulative edge length, thereby effectively approximating the skeletal structure of the underlying road network. For any given bus location and its designated target stop, the optimal shortest path $\mathcal{P}'$ is then computed on this MST-augmented graph. This computation employs Dijkstra's algorithm to identify the trajectory that minimises the total geographic distance traversed. Crucially, this derived path $\mathcal{P}'$ precisely defines the expected route sequence, ensuring consistency across all subsequent arrival time estimation models and underpinning the logical coherence of the prediction process.

\subsection{ETA Prediction Models}
To accurately estimate travel time along the precisely computed path $\mathcal{P}'$, this study presents and carefully analyses six distinct modelling approaches. This deliberate multi-model strategy facilitates a comprehensive and nuanced evaluation of various theoretical paradigms, ranging from straightforward historical averaging techniques to sophisticated and complex hybrid deep learning architectures. Such an extensive comparative analysis is critical for discerning the most robust and accurate predictive mechanism under diverse operational conditions and for understanding the relative strengths and limitations of each methodological class.
\subsubsection{Model 1: MST-AV (Historical Average Baseline)}
The MST with average velocity model constitutes a fundamental, non-temporal historical baseline, offering a foundational reference point for travel time prediction. Its theoretical underpinning relies on the premise that future vehicular speeds can be reliably extrapolated from observed historical averages and a widely recognized approach in various forecasting domains. This model establishes a critical performance floor against which the efficacy and incremental value of more sophisticated predictive models must be rigorously benchmarked and validated. (See Training: Algorithm~\ref{alg:mst_av_training} and Prediction: Algorithm~\ref{alg:mst_av_computation}) \\

Initially, it involves computing the mean historical speed, denoted as $\mathcal{\widehat{U}}_i^{\text{avg}}$, for each node $i$ in the network by aggregating all historical speed observations at that location.\\

For prediction, the model calculates the estimated travel time for each constituent edge $(i,j)$ within the designated path $\mathcal{P}'$ by leveraging the mean historical speed observed at its connected vertices. Specifically, the average velocity for an edge, $\mathcal{\widehat{U}}_{ij}^{\text{avg}}$, is determined as the arithmetic mean of the historical average speeds of its incident nodes:

\[
    \mathcal{\widehat{U}}_{ij}^{\text{avg}} = \frac{\mathcal{\widehat{U}}_i^{\text{avg}} + \mathcal{\widehat{U}}_j^{\text{avg}}}{2}
\]

Consequently, the total Estimated Time of Arrival ($\mathcal{ETA}_{\text{MST-AV}}$) for the entire path is precisely formulated as:

\[
\mathcal{ETA}_{\text{MST-AV}} = \sum_{(i,j) \in \mathcal{P}'} \frac{d(i,j)}{\mathcal{\widehat{U}}_{ij}^{\text{avg}}}
\]

where $d(i,j)$ denotes the Haversine distance separating nodes $i$ and $j$.

\subsubsection{Model 2: GDRN-DFT (Spectral Pattern Analysis)}
\begin{figure}[htbp]
    \centering
    \includegraphics[width=1\textwidth]{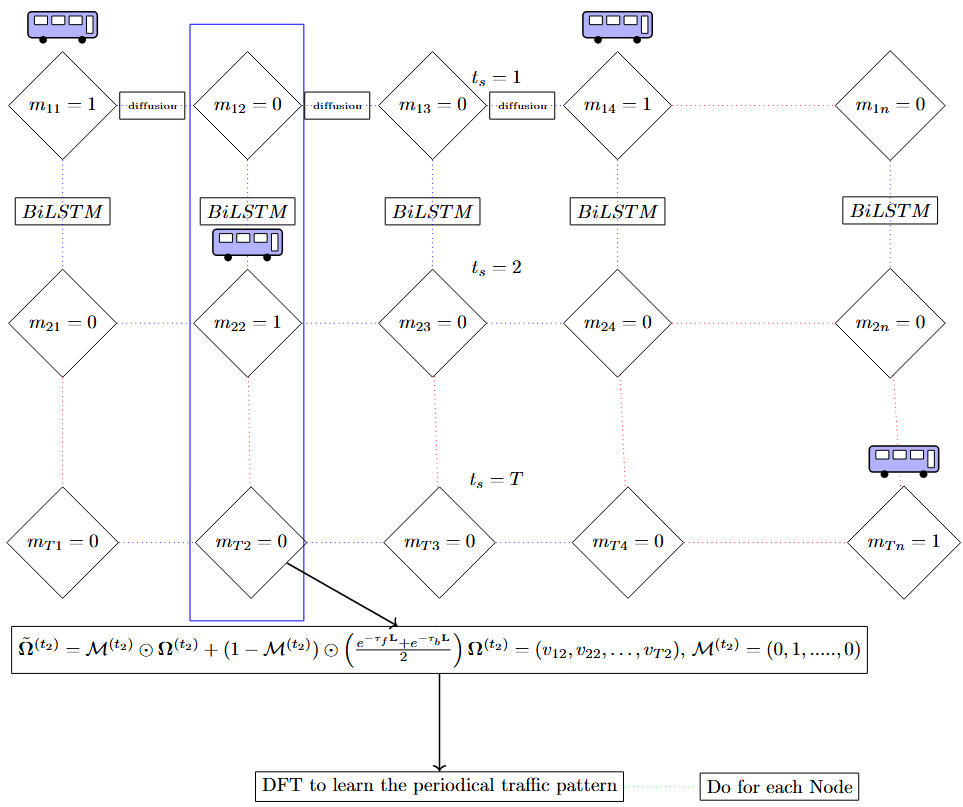}
    \caption{GDRN-DFT Diagram.(For training : Algorithm~\ref{alg:gdrn_dft_training} and Computation of real time ETA : Algorithm ~\ref{alg:gdrn_dft_computation})}
    \label{fig:fig_gdrndft}
\end{figure}

To precisely capture and model inherent periodic traffic patterns, such as distinct daily and weekly cycles, the present methodology integrates a Graph Diffusion Recurrent Network (see Fig.~\ref{fig:fig_gdrndft}). This advanced framework is specifically engineered for the spatiotemporal imputation of missing speed values across the entire network, serving as an essential prerequisite for subsequent spectral analysis.\\

The GDRN leverages graph diffusion mechanisms\cite{10508504} for effective spatial data smoothing. Initially, the normalized graph Laplacian $\mathbf{L} = \mathbf{I} - \mathbf{D}^{-1/2} \mathbf{A} \mathbf{D}^{-1/2}$ (where $\mathbf{A}$ is the adjacency matrix and $\mathbf{D}$ is the degree matrix) is computed to represent the network topology. A diffused signal $\tilde{\mathbf{\Omega}}^{(t_s)}$ at time $t_s$ is then generated, which smooths the observed speeds across connected nodes. Then the system performs an intelligent imputation step using a mask $\mathbf{\mathcal{M}}^{(t_s)}$, which identifies observed data points. Missing values are filled with their diffused counterparts to form $\tilde{\mathbf{\Omega}}^{(t_s)}$:

\[
\tilde{\mathbf{\Omega}}^{(t_s)} = \mathbf{\mathcal{M}}^{(t_s)} \odot \mathbf{\Omega}^{(t_s)} + (1 - \mathbf{\mathcal{M}}^{(t_s)}) \odot \left( \frac{e^{-\tau_f \mathbf{L}} + e^{-\tau_b \mathbf{L}}}{2} \right) \mathbf{\Omega}^{(t_s)}
\]

Subsequently, the framework incorporates a bidirectional Long Short-Term Memory (bi-LSTM)\cite{9005997} component to robustly model intricate temporal dependencies within the imputed speed series. The forward and backward LSTM passes ($\overrightarrow{\mathbf{h}}_{t_s}$ and $\overleftarrow{\mathbf{h}}_{t_s}$) process the diffused and imputed speeds, and their hidden states are concatenated ($\mathbf{h}_{t_s} = [\overrightarrow{\mathbf{h}}_{t_s}; \overleftarrow{\mathbf{h}}_{t_s}]$) to capture both past and future contextual information. The final predicted speed vector $\hat{\mathbf{\Omega}}^{(t_s)}$ is then generated through a linear transformation of this concatenated hidden state:

\[
\hat{\mathbf{\Omega}}^{(t_s)} = \mathbf{W}_h \mathbf{h}_{t_s} + \mathbf{b}_h
\]

The entire GDRN is carefully optimized through training with a specifically formulated loss function, $\mathcal{L}_{gd}$, which ensures accurate imputation and prediction by focusing solely on the discrepancies at observed locations:

\[
\mathcal{L}_{gd} = \left| \mathbf{\mathcal{M}}^{(t_s)} \odot (\mathbf{\Omega}^{(t_s)} - \hat{\mathbf{\Omega}}^{(t_s)}) \right|_2^2
\]

This rigorous optimization process facilitates the creation of a comprehensive, imputed speed time series for each node, which is then prepared for Discrete Fourier Transform (DFT) analysis. The DFT\cite{gurevich2008discretefouriertransformcanonical} step, subsequently applied after this training process to the full imputed dataset, accurately identifies dominant periodic components, a methodological choice theoretically justified by the strong, demonstrably cyclical nature of urban traffic dynamics. Ultimately, the application of an inverse DFT provides smoothed, forecasted speed values for each node and future timestep, which are then critically utilised for precise, time-dependent ETA calculations.

\subsubsection{Model 3: KNN (Temporal Dynamics via Koopman Theory)}

\begin{figure}[htbp]
    \centering
    \includegraphics[width=1\textwidth]{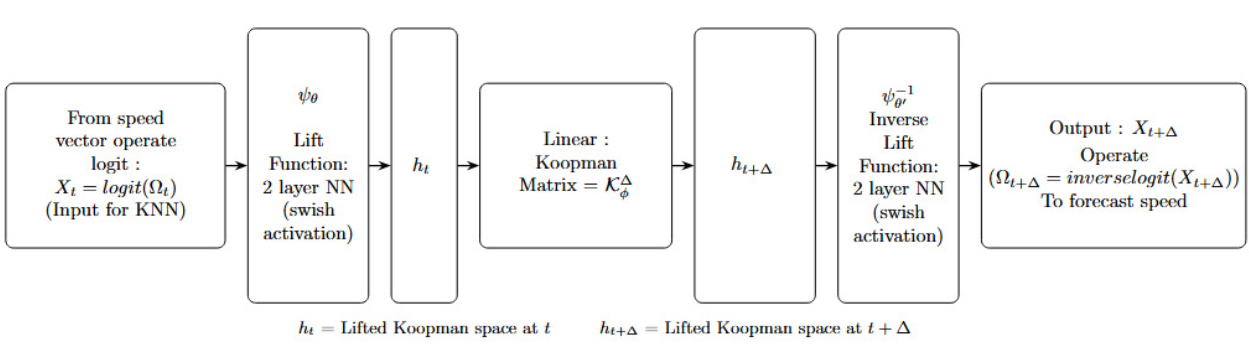}
    \caption{KNN Diagram. (see for training : Algorithm~\ref{alg:knn_training}  and  real time ETA computation : Algorithm~\ref{alg:knn_computation} .)}
    \label{fig:Knn_diagram}
\end{figure}

To rigorously model and predict the inherently nonlinear temporal dynamics of bus speeds, a sophisticated Koopman Neural Network (KNN) \cite{dogra2020optimizingneuralnetworkskoopman} is strategically deployed. Its foundational theoretical underpinnings are rooted in Koopman operator theory, which elegantly posits that inherently nonlinear dynamic systems can be precisely represented through the action of a linear operator within an infinite-dimensional function space of observables. (See Figure.~\ref{fig:Knn_diagram})\\

Practically, this is realised by an encoder network, $\psi_\theta$, which adeptly learns a finite-dimensional lifting of the nonlinear speed time series. This process transforms an input sequence of logit-transformed speeds, $\mathbf{x}_t = [\omega_{t-(k-1)}, \dots, \omega_t]$, into a latent, embedded space where the system's dynamics linearly evolve. The logit transformation, $\omega_j = \ln(v'_j / (V_{\max} - v'_j))$, is applied to filter the raw speeds $v'_j$ (derived from historical average speeds. $v'_j = \min(V_{\max} - \epsilon, \max(v_j, \epsilon))$), enhancing numerical stability and boundedness with in $[0, V{\max}]$. The encoder maps this input to a latent representation $h_t$:

\[
h_t = \psi_\theta(\mathbf{x}_t)
\]

In this latent space, a Koopman matrix $\mathcal{K}_\phi \in \mathbb{R}^{D_k \times D_k}$ linearly approximates the evolution of the system. For multi-step prediction, the latent state at $\Delta$ future time steps, $h_{t+\Delta}$, is approximated by recursively applying the Koopman operator:

\[
h_{t+\Delta} = \mathcal{K}_\phi^\Delta h_t
\]

Subsequently, a corresponding decoder component, $\psi_{\theta'}^{-1}$, facilitates the inverse mapping, projecting the linear predictions from the latent space back into the original, observable data space. This reconstruction yields both the initial state $\hat{\mathbf{x}}_t$ from $h_t$ and future predictions $\hat{\mathbf{x}}_{t+\Delta}$ from $h_{t+\Delta}$:

\[
\hat{\mathbf{x}}_t = \psi_{\theta'}^{-1}(h_t) \quad \text{and} \quad \hat{\mathbf{x}}_{t+\Delta} = \psi_{\theta'}^{-1}(h_{t+\Delta})
\]

This sophisticated methodology proves exceptionally efficacious for robust multi-step forecasting, while simultaneously furnishing a powerful and coherent theoretical framework for data-driven learning of evolution operators from empirical observations. The KNN is trained by minimizing a composite loss function,$\mathcal{L}_{\text{KNN}}$, which balances the accuracy of state reconstruction with the fidelity of future predictions:

\[
\mathcal{L}_{\text{KNN}} = |\mathbf{x}_t - \hat{\mathbf{x}}_t|^2 + \lambda_{knn} \sum_\Delta |\mathbf{x}_{t+\Delta} - \hat{\mathbf{x}}_{t+\Delta}|^2
\]

Where, the first term represents the reconstruction loss, ensuring the encoder-decoder pair accurately captures the input state, and the second term, weighted by the hyperparameter $\lambda_{knn}$, represents the prediction loss across various future time steps $\Delta$, driving the model to accurately forecast future speed dynamics.
\subsubsection{Model 4: FENN (Feature-Engineered Forecasting)}

The Feature-Encoded Neural Network (FENN) represents a significant methodological advancement by explicitly integrating a rich array of exogenous contextual factors into the speed prediction process, thereby transcending the limitations inherent in pure time-series models. For each discrete timestamp $t$, a comprehensive feature vector $\mathbf{y}_t$ is meticulously constructed. This vector systematically incorporates crucial contextual elements, including precise geospatial coordinates $(\vartheta_i, \varphi_i)$, dynamic weather data $\mathcal{W}_i$ retrieved via API, and $t$ This temporal feature is ingeniously encoded as sinusoidal functions, $\varsigma_i(m) = \sin(\alpha_i + 2m\pi / b)$ for $m=0,\dots,b-1$ (where $\alpha_i = 2\pi / 24 \cdot \text{hour}(t_i)$), to robustly capture cyclical patterns in a day. The complete feature vector for a given time $i$ and its associated speed $v_i$ is defined as:

\[
\mathbf{y_i} = (\vartheta_i, \varphi_i, \mathcal{W}_i, \varsigma_i, \dots, \varsigma_i(b-1), v_i)
\]

For prediction, a sequence of these $\mathbf{y_i}$ vectors, spanning a defined window $k$, forms an input matrix :\\ $\mathbf{Y_t} = (\mathbf{y}_{t-(k-1)}, \dots, \mathbf{y_t})$. Each individual feature vector $\mathbf{y_i}$ within this sequence is then processed by a neural network encoder, which compresses this rich, high-dimensional input into a more compact and meaningful representation, $\zeta_i$. These encoded representations collectively form the input $\mathbf{Z_t} = (\zeta_{t-(k-1)}, \dots, \zeta_t)$ for the subsequent predictive network.\\ 

To facilitate more effective learning for speed values, the target values are transformed using a logit function. For future speed values $v_j$, the transformed target $\omega_j$ is computed as $\omega_j = \ln(v'_j / (V_{\max} - v'_j))$, where $V_{\max}$ is the maximum possible speed and $v'_j = \min(V_{\max} - \epsilon, \max(v_j, \epsilon))$. These transformed targets for a future window are aggregated into $\mathbf{X_{t+\Delta}}$. A dedicated feedforward neural network, $\mathcal{F}_{\text{NN}}$, then leverages the encoded information $\mathbf{Z_t}$ to forecast these future transformed speed values, denoted as $\hat{\mathbf{X}}_{t+\Delta}$.(See for Training : Algorithm~\ref{alg:fenn_training} and for real time ETA Algorithm~\ref{alg:fenn_computation} for real time ETA computation)\\

The FENN model is optimised through training, minimising the sum of squared errors between the actual transformed speeds $\mathbf{X_{t+\Delta}}$ and the predicted transformed speeds $\hat{\mathbf{X}}_{t+\Delta}$. This objective is formally expressed by the loss function $\mathcal{L}_{\text{FN}}$:

\[
\mathcal{L}_{\text{FN}} = \sum_{\Delta} |\mathbf{X_{t+\Delta}} - \hat{\mathbf{X}}_{t+\Delta}|_2^2
\]
\subsubsection{Model 5: MGCN(Real-Time Graph Learning)}

The Masked Graph Convolutional Network \cite{yang2019masked} is a sophisticated architecture strategically designed for real-time traffic speed prediction and imputation within an intricate transportation network. Its primary function is the dynamic fusion of currently observed speeds with prior, often complementary, speed estimates—typically derived from methods like Discrete Fourier Transform. This synergistic integration is managed through a dynamic binary mask $\mathcal{M}$, which precisely indicates the availability of real-time observations for each node at a given timestamp.
\begin{figure}[htbp]
    \centering
    \includegraphics[width=1\textwidth]{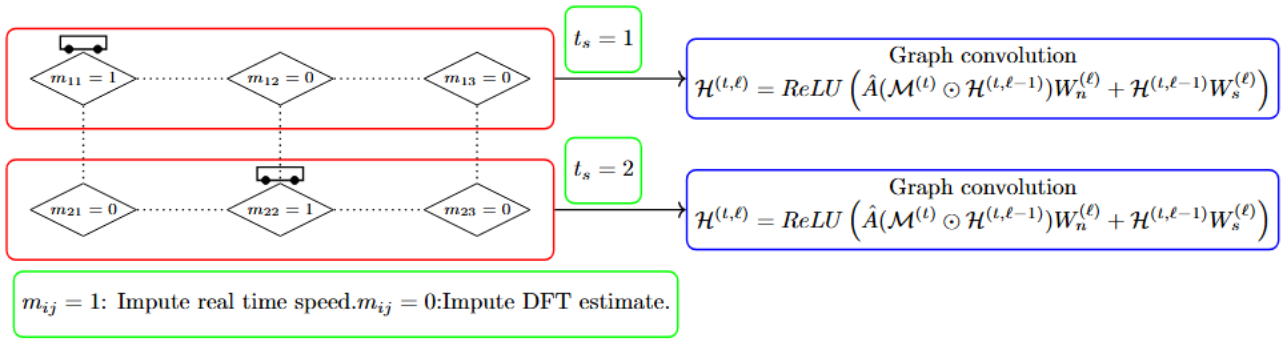}
    \caption{MGCN diagram.(Algorithm~\ref{alg:mgcn_training} for training and  Algorithm~\ref{alg:mgcn_computation} for real time ETA computation.)}
    \label{fig:fig_mgcn+}
\end{figure}
At its core, the MGCN leverages Graph Convolutional Networks to intelligently propagate spatiotemporal information across the network's topological structure. This propagation is crucial for adeptly imputing data for unobserved nodes (where $\mathcal{M}_{i}^{(t)} = 0$) while simultaneously preserving the integrity and signal fidelity of speeds from nodes where real-time observations are available (where $\mathcal{M}_{i}^{(t)} = 1$). The input to the GCN layers at each time step $t$ is a hybrid feature matrix $\mathcal{H}^{(t)}$, formed by combining observed speeds $\Omega_{\text{obs}}^{(t)}$ and DFT speeds $\Omega_{\text{dft}}^{(t)}$ using the mask $\mathcal{M}^{(t)}$:

\[
\mathcal{H}^{(t)} = \mathcal{M}^{(t)} \odot \Omega_{\text{obs}}^{(t)} + (1 - \mathcal{M}^{(t)}) \odot \Omega_{\text{dft}}^{(t)}
\]
The GCN's $L$ layers then process this hybrid input, with each layer $\ell$ performing a graph convolution and non-linear activation:

\[
\mathcal{H}^{(t,\ell)} = \text{ReLU}\left( \hat{A} (\mathcal{M}^{(t)} \odot \mathcal{H}^{(t,\ell-1)}) W_n^{(\ell)} + \mathcal{H}^{(t,\ell-1)} W_s^{(\ell)} \right)
\]

Where $\hat{A}$ represents the normalised adjacency matrix of the transportation graph, and $W_n^{(\ell)}$, $W_s^{(\ell)}$ are learnable weight matrices for the neighbourhood and self-connections, respectively. This iterative process culminates in a predicted speed map $\hat{\Omega}^{(t+1)}$ for the subsequent time step.\\

The entire MGCN model is meticulously optimised through training with a specifically formulated composite loss function, $\mathcal{L}_{gcn}$, which ensures robust performance by balancing accuracy on observed data with consistency against prior estimates for unobserved data. 

\[
\mathcal{L}_{gcn} = \frac{1}{\sum \mathcal{M}^{(t+1)}} \sum \mathcal{M}^{(t+1)} (\hat{\Omega}^{(t+1)} - \Omega_{\text{obs}}^{(t+1)})^2
\]
\\
where $\mathcal{L}_{gcn}$ quantifies the mean squared error only for observed speeds, ensuring high fidelity where data is available:

This rigorous optimization process ultimately yields a highly robust and comprehensive real-time speed map.

\subsubsection{Model 6: HYB($n_m$)(Hybrid Framework)}

\begin{figure}[htbp]
    \centering
    \includegraphics[width=.8\textwidth]{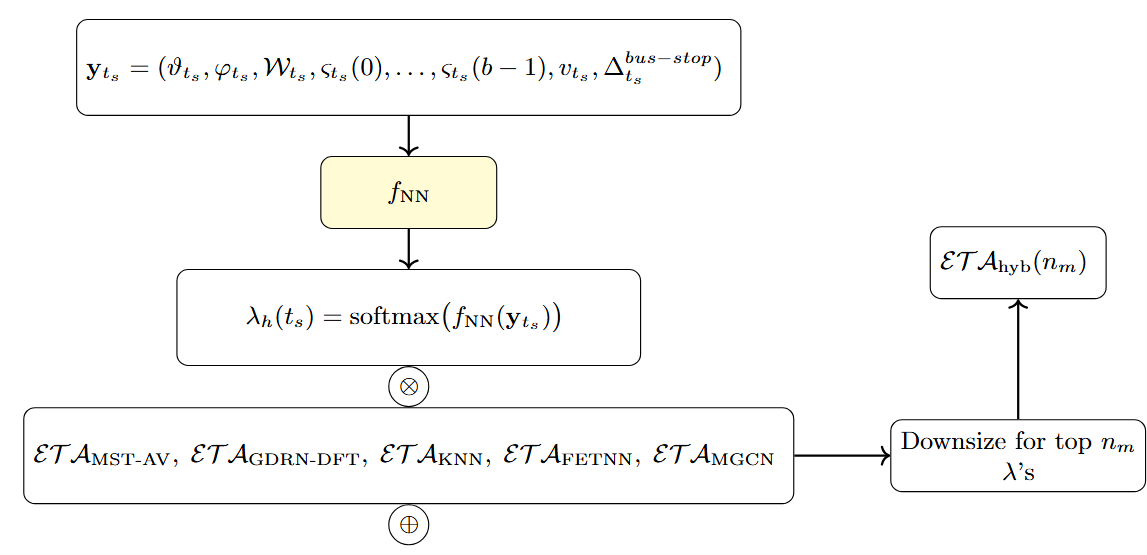}
    \caption{Hybrid ETA diagram.(Algorithm ~\ref{alg:hyb_nm_training} for training and Algorithm~\ref{alg:hyb_nm_computation} for real time ETA computation)}
    \label{fig:fig_hyb_eta}
\end{figure}

 The finale framework introduces an adaptive hybrid ensemble meticulously engineered to integrate and synergistically fuse predictions from a set of five distinct base models. The fundamental premise of this design acknowledges that the optimal predictive model is inherently dynamic, varying with prevailing contextual conditions. (See figure ~\ref{fig:fig_hyb_eta} to understand the hybridization mechanism.)\\
To effectuate this dynamic integration, a meta-learner neural network, denoted as $f_{\text{NN}}$, is employed to generate adaptive blending weights. These weights, $\lambda_{ih}$, where $i \in {1, \dots, 5}$ signifies a base model and $h$ an instance, are determined by mapping a comprehensive feature vector, $\mathbf{y}_{t_s}$, observed at timestamp $t_s$. The feature vector $\mathbf{y}_{t_s}$ encapsulates:

\[
\mathbf{y}_{t_s} = (\vartheta_{t_s}, \varphi_{t_s}, \mathcal{W}_{t_s}, \varsigma_{t_s}(0), \dots, \varsigma_{t_s}(b-1), v_{t_s}, \Delta_{t_s}^{bus-stop})
\]
Where, $(\vartheta_{t_s}, \varphi_{t_s})$ represent geospatial coordinates, $\mathcal{W}_{t_s}$ denotes environmental weather data, $\varsigma_{t_s}(m) = \sin(\alpha_{t_s} + 2m\pi / b)$ for $m=0,\dots,b-1$ (with $\alpha_{t_s} = 2\pi / 24 \cdot \text{hour}(t_s)$) captures periodic temporal dependencies, $v_{t_s}$ is the current vehicle speed, and $\Delta_{t_s}^{bus-stop}$ represents the distance computed via graph from the current location to the target. The raw blending weights are thus the output of the meta-learner: $\lambda_h(t_s) = softmax(f_{\text{NN}}(\mathbf{y}_{t_s}))$.\\

To enhance computational efficiency and interpretability, a sophisticated downsizing mechanism is integrated, controlled by a hyperparameter $n_m \le 5$. This mechanism selectively activates only the $n_m$ models assigned the highest raw blending weights by $f_{\text{NN}}$. A binary mask $m_i$ is constructed, where $m_i=1$ if model $i$ is among the top-$n_m$ and $m_i=0$ otherwise. The final, normalized blending weights, $\lambda'_{ih}$, are then computed as:

\[
\lambda'_{ih}(t_s) = \frac{m_i \cdot \lambda_{ih}(t_s)}{\sum_{j=1}^5 m_j \cdot \lambda_{jh}(t_s)}
\]

The ensemble's estimated time of arrival, $\mathcal{ETA}'(n_m)_{\text{hyb}}$, is subsequently derived as the weighted sum of the individual model ETAs, $\mathcal{ETA}_i$:

\[
\mathcal{ETA}'(n_m)_{\text{hyb}} = \sum_{i=1}^5 \lambda'_{ih} \cdot \mathcal{ETA}_i
\]

The meta-learner $f_{\text{NN}}$ is trained by minimizing a loss function: $\mathcal{L}_{\text{hyb}}$, which combines a constraint on weight summation and the squared error between the ensemble prediction and the actual ETA:

\[
\mathcal{L}_{\text{hyb}} = \left(\sum_{j=1}^5\lambda_{jh}(t_s).\mathcal{ETA}_j - \mathcal{ETA}_{\text{actual}}\right)^2
\]

\section{Result and Discussion}
This section presents a comprehensive evaluation of the proposed individual models developed for bus arrival time prediction across three distinct routes (1, 2, and 3).
In addition to analyzing their individual performance, the work also investigates the efficacy of their hybridization, conducts a thorough sensitivity analysis for various hyperparameter settings, and investigates the effects of downsizing within the hybrid model. Last, work findings are benchmarked against established deep learning architectures to provide a robust comparative analysis.

\subsection{Evaluation Metrics}

In this work, three widely used metrics such as Mean Absolute Error, Root Mean Square Error, and Mean Absolute Percentage Error are used as model performance metrics which provide a holistic view of predictive accuracy. The individual metrics definitions are as follows:

\textbf{Mean Absolute Error (MAE):} Measures the average magnitude of the errors, quantifying the difference between predicted and actual values in minutes.
\[
\text{MAE} = \frac{1}{N} \sum_{i=1}^N |\mathcal{ETA}_{\text{pred},i} - \mathcal{ETA}_{\text{actual},i}| \quad (\text{minutes})
\]

\textbf{Root Mean Square Error (RMSE):} Measures the average magnitude of the errors, with a greater penalty for larger errors due to the squaring term in minutes.
\[
\text{RMSE} = \sqrt{\frac{1}{N} \sum_{i=1}^N (\mathcal{ETA}_{\text{pred},i} - \mathcal{ETA}_{\text{actual},i})^2} \quad (\text{minutes})
\]

\textbf{Mean Absolute Percentage Error (MAPE):} expresses the accuracy as a percentage of the actual values, making it scale-independent and particularly useful for comparing models across different contexts.
\[
\text{MAPE} = \frac{100}{N} \sum_{i=1}^N \left| \frac{\mathcal{ETA}_{\text{pred},i} - \mathcal{ETA}_{\text{actual},i}}{\mathcal{ETA}_{\text{actual},i}} \right| \%
\] 

All evaluations were systematically conducted on a dedicated test set comprising $N=10,000$ bus trips per route, utilizing GPS-derived ground-truth arrival times for accurate assessment.

\subsection{Performance Comparison Between the Models}
Model performance values of  MAE, RMSE, and MAPE for route 1 are tabulated in Table~\ref{tab:model_performance_individual}. Furthermore, model performance is analysed under following several challenging scenarios to better understand the model's robustness and applicability:\\

\begin{description}
    \item[\textbf{Normal Condition:}] Investigated individual model performance as a function of varying actual Estimated Time of Arrival ranges, assessing accuracy across short, medium, and long prediction windows.\\
    \item[\textbf{Prediction Horizon:}] The efficacy of each individual model is examined based on the prediction horizon, specifically evaluating accuracy for upcoming stop levels, from the immediate next stop to multiple subsequent stops.\\
    \item[\textbf{Extreme Weather Conditions:}] Individual model resilience was tested under challenging environmental conditions, including periods of heavy rain, dense cloud cover, and fog, to ascertain their robustness against real-world fluctuations.
\end{description}
\begin{table}[h!]
\centering
\setlength{\tabcolsep}{2.9pt}
\caption{Route 1: Model performance values of MAE, MAPE \%, RMSE of individual and hybridization models for route-1. }
\label{tab:model_performance_individual}
\fontsize{5}{5}\selectfont
\begin{tabular}{@{}lccccccccccccccccc@{}}
\toprule
 & & \multicolumn{5}{c}{Normal Conditions} & \multicolumn{4}{c}{Bus Stop Groups} & \multicolumn{5}{c}{Extreme Weather} \\
 & & \multicolumn{5}{c}{(Actual ETA ranges(min))} & \multicolumn{4}{c}{} & \multicolumn{5}{c}{(Actual ETA ranges(min))} \\
\cmidrule(r){3-7} \cmidrule(lr){8-11} \cmidrule(l){12-16}
Model & Metric & 0-10 & 10-25 & 25-45 & 45+ & Overall & 1-2 & 3-4 & 5-6 & 7+ & 0-10 & 10-25 & 25-45 & 45+ & Overall \\
\midrule
MST-AV & MAE  & 4.2 & 5.9 & 8.2 & 10.3 & 8.0 & 4.7 & 6.8 & 8.9 & 11.1 & 6.0 & 7.2 & 8.9 & 11.4 & 8.8 \\
 & MAPE & 60.9 & 35.7 & 25.6 & 17.4 & 21.1 & 44.8 & 28.9 & 20.4 & 18.2 & 75.9 & 40.1 & 28.7 & 20.4 & 25.1 \\
 & RMSE & 5.4 & 7.2 & 9.8 & 12.7 & 9.6 & 6.0 & 8.4 & 10.7 & 13.6 & 7.1 & 8.6 & 10.4 & 13.1 & 10.2 \\
\midrule
GDRN-DFT & MAE  & 3.4 & 5.6 & \textbf{6.1} & \textbf{7.2} & 5.9 & 4.1 & 6.9 & \textbf{7.3} & \textbf{8.6} & 5.4 & 7.0 & 7.9 & \textbf{8.8} & 8.0 \\
 & MAPE & 42.5 & 31.1 & \textbf{17.7} & \textbf{11.4} & 14.6 & 36.6 & 27.6 & \textbf{16.7} & \textbf{14.2} & 63.5 & 33.3 & 22.6 & \textbf{14.7} & 21.3 \\
 & RMSE & 4.5 & 6.8 & 7.7 & \textbf{9.1} & 7.4 & 5.3 & 8.5 & \textbf{8.9} & \textbf{10.2} & 6.3 & 8.2 & 9.1 & \textbf{10.2} & 9.0 \\
\midrule
KOOP-NET & MAE  & \textbf{1.2} & 4.5 & 9.2 & 11.8 & 8.8 & \textbf{2.3} & 5.2 & 10.6 & 13.1 & 3.1 & 5.8 & 7.8 & 10.2 & 8.6 \\
 & MAPE & \textbf{17.9} & 29.5 & 29.2 & 20.1 & 23.7 & \textbf{20.3} & 22.8 & 25.5 & 22.6 & 42.5 & 30.5 & 22.9 & 18.3 & 23.6 \\
 & RMSE & \textbf{1.9} & 6.0 & 11.0 & 14.3 & 10.5 & 3.2 & 6.7 & 12.8 & 15.7 & 3.9 & 6.9 & 9.1 & 11.7 & 9.5 \\
\midrule
FE-NN & MAE  & 2.8 & 4.1 & 8.6 & 10.6 & 8.3 & 2.5 & 5.0 & 10.2 & 12.3 & \textbf{2.2} & \textbf{4.3} & \textbf{6.6} & 9.1 & 6.7 \\
 & MAPE & 37.3 & 27.2 & 26.1 & 17.4 & 21.3 & 23.2 & 20.7 & 22.2 & 19.7 & \textbf{24.7} & 26.9 & \textbf{20.3} & 15.7 & 19.3 \\
 & RMSE & 3.9 & 5.4 & 10.2 & 12.8 & 9.9 & 3.4 & 6.4 & 12.3 & 14.8 & \textbf{3.0} & \textbf{5.4} & \textbf{7.7} & 10.6 & 8.1 \\
\midrule
RT-GCN & MAE  & 2.6 & \textbf{3.7} & 8.9 & 12.9 & 7.4 & 2.6 & \textbf{4.7} & 10.1 & 13.8 & 2.9 & 4.9 & 7.2 & 10.8 & 7.4 \\
 & MAPE & 34.9 & \textbf{21.6} & 28.5 & 22.0 & 19.9 & 25.2 & \textbf{18.4} & 22.1 & 23.4 & 42.0 & 26.9 & 21.6 & 19.4 & 21.0 \\
 & RMSE & 3.5 & \textbf{4.9} & 10.5 & 15.6 & 9.3 & 3.6 & \textbf{6.0} & 12.6 & 16.5 & 3.7 & 6.1 & 8.5 & 12.5 & 8.9 \\
\midrule
HYB-COMB\\(5) & MAE  & 1.5 & 3.9 & 6.2 & 7.6 & \textbf{4.8} & 2.4 & 4.9 & 7.8 & 8.9 & 2.4 & 4.6 & 6.9 & 8.9 & \textbf{6.1} \\
 & MAPE & 22.8 & 22.0 & 17.9 & 12.0 & \textbf{13.7} & 20.5 & 19.2 & 17.1 & 14.4 & 26.7 & \textbf{23.1} & 20.8 & 14.9 & \textbf{18.1} \\
 & RMSE & 2.2 & 5.0 & \textbf{7.6} & 9.3 & \textbf{6.5} & \textbf{3.1} & 6.2 & 9.5 & {10.8} & 3.2 & 5.6 & {8.0} & {10.4} & \textbf{7.6} \\
\bottomrule
\end{tabular}
\end{table}

As detailed in Table \ref{tab:model_performance_individual} and figure ~\ref{fig:indv_analysis}, (Table ~\ref{tab:model_performance_individual} for route 1. for Route 2 and Route 3 see Appendix. - Tables: ~\ref{tab:performance_study_route2},~\ref{tab:performance_study_route3}) our comprehensive evaluation of bus arrival time prediction models reveals nuanced performance characteristics across various conditions. 
\begin{figure}[htbp]
    \centering
    \includegraphics[width=1\linewidth]{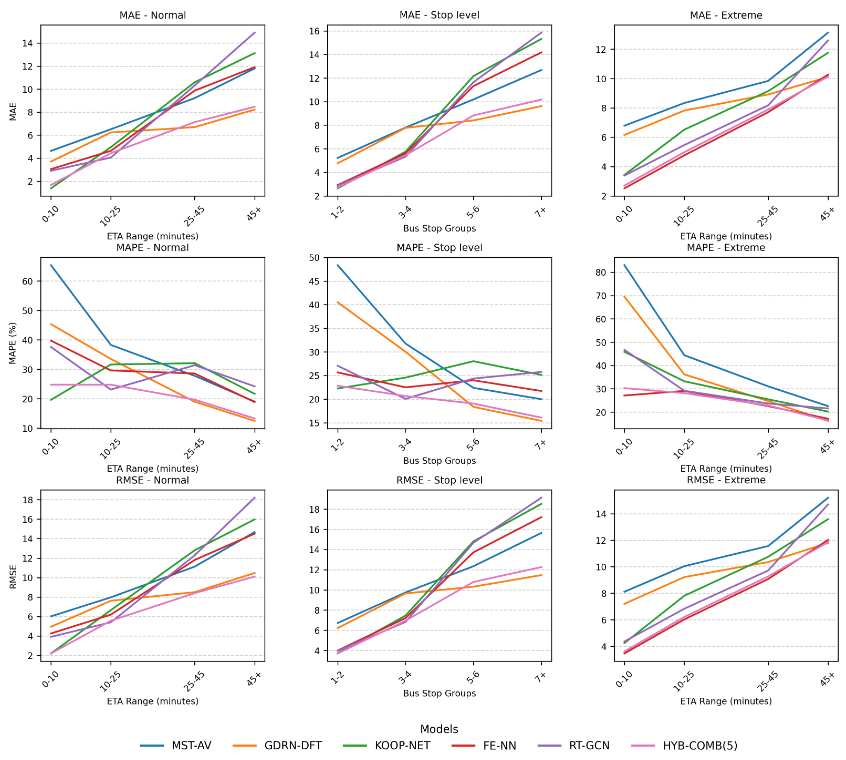}
    \caption{The figure displays a 3x3 grid of line plots showing the average performance (MAE, MAPE, RMSE) of six models across Routes 1, 2, and 3 for Normal Conditions, Bus Stop Groups, and Extreme Weather, highlighting fluctuations across ETA ranges and bus stop groups.}
    \label{fig:indv_analysis}
\end{figure}
The HYB-COMB model consistently demonstrated superior overall accuracy, exhibiting the lowest MAE, MAPE, and RMSE across normal, bus stop group, and extreme weather scenarios, thereby establishing itself as the most robust and reliable predictor. For short-range predictions (0-10 minutes or 1-2 bus stops), KOOP-NET proved exceptionally accurate under normal conditions, while FE-NN excelled in the same range during extreme weather, showcasing specialised strengths. RT-GCN delivered competitive results for medium distances (10-25 minutes or 3-4 bus stops), particularly under normal conditions. However, GDRN-DFT emerged as the dominant model for longer prediction horizons (25-45 and 45+ minutes, or 5-6 and 7+ bus stops), consistently achieving the lowest errors across all conditions in these extended ranges. Conversely, the MST-AV model consistently yielded the highest errors, indicating its limitations, while KOOP-NET, despite its initial strengths, struggled with scalability to longer distances. Notably, all models generally experienced increased prediction errors during extreme weather conditions, underscoring the inherent challenges of forecasting under adverse environmental factors. In essence, while specialised models offer targeted accuracy, HYB-COMB provides the most balanced and resilient performance for general application.\\

\subsection{Downsizing Effect of Hybrid Model}
This experimental investigation meticulously evaluates the predictive capabilities of five distinct hybrid models, designated HYB($n_m$) through HYB(5), for estimating bus arrival times. Our analysis spans three heterogeneous routes (Routes 1, 2, and 3) and considers various prediction horizons, categorised into four crucial Estimated Time of Arrival ranges: 0-10, 10-25, 25-45, and 45+ minutes. Central to this approach is the HYB($n_m$) framework, which ingeniously employs a meta-learner neural network. This sophisticated component adaptively synthesises predictions from five foundational base models, leveraging a rich array of contextual features including precise geospatial coordinates, prevailing weather conditions, and real-time vehicle speed. The efficacy of these hybrid models is rigorously quantified using widely recognised performance metrics: MAE, MAPE, and RMSE.

\begin{table}[h!]
\centering
\setlength{\tabcolsep}{0pt}
\caption{Comparison of MAE, MAPE (\%), and RMSE for HYB models across ETA ranges in Route 1, Route 2, and Route 3.(Lowest values per route are in bold.)}
\label{tab:hyb-sidebyside}
\fontsize{5.0}{5.0}\selectfont
\renewcommand{\arraystretch}{1.10}
\setlength{\tabcolsep}{1.3pt} 
\begin{tabular}{l l | @{\hskip 1pt}ccccc@{\hskip 1pt} | @{\hskip 1pt}ccccc@{\hskip 1pt} | @{\hskip 1pt}ccccc@{\hskip 1pt}}
\toprule
\multirow{2}{*}{\textbf{Model}} & \multirow{2}{*}{\textbf{Metric}} 
& \multicolumn{5}{c}{\textbf{Route 1}} 
& \multicolumn{5}{c}{\textbf{Route 2}} 
& \multicolumn{5}{c}{\textbf{Route 3}} \\
\cmidrule(lr){3-7} \cmidrule(lr){8-12} \cmidrule(lr){13-17}
& & 0--10 & 10--25 & 25--45 & 45+ & Overall 
& 0--10 & 10--25 & 25--45 & 45+ & Overall 
& 0--10 & 10--25 & 25--45 & 45+ & Overall \\
\midrule
HYB(1) & MAE  & {1.4} & 4.1 & \textbf{6.2} & 7.4 & \textbf{4.3}
                 & \textbf{1.6} & 5.0 & \textbf{8.0} & 9.1 & \textbf{5.4}
                 & \textbf{1.5} & 4.6 & \textbf{7.2} & 8.4 & {5.0} \\
       & MAPE & 20.5 & 24.2 & 18.8 & \textbf{10.9} & 13.2
                 & 24.4 & 31.3 & 22.4 & \textbf{13.0} & 15.9
                 & 22.9 & 29.0 & 20.5 & \textbf{12.3} & 14.7 \\
       & RMSE & 1.8 & 5.3 & 7.6 & {8.5} & 5.6
                 & 2.2 & 6.5 & 9.5 & \textbf{10.6} & 7.1
                 & \textbf{2.0} & 6.0 & 8.7 & \textbf{9.7} & \textbf{6.1} \\
\midrule
HYB(2)\\Proposed Model & MAE  & \textbf{1.3} & 4.3 & 6.5 & \textbf{7.3} & 4.5
                 & 1.7 & 5.3 & 8.1 & \textbf{9.0} & 5.7
                 & 1.6 & 4.8 & 7.4 & \textbf{8.3} & \textbf{4.8} \\
       & MAPE & \textbf{19.4} & 23.5 & \textbf{17.8} & 11.2 & \textbf{12.3}
                 & \textbf{23.1} & 28.0 & \textbf{21.2} & 13.3 & \textbf{14.9}
                 & \textbf{21.7} & 25.9 & \textbf{19.6} & 12.6 & \textbf{13.8} \\
       & RMSE & \textbf{1.7} & 5.0 & \textbf{7.3} & \textbf{8.4} & \textbf{5.4}
                 & \textbf{2.0} & 6.1 & \textbf{9.1} & 10.7 & \textbf{6.9}
                 & {2.1} & 5.7 & \textbf{8.4} & 9.8 & {6.4} \\
\midrule
HYB(3) & MAE  & 1.6 & \textbf{3.8} & 6.4 & 7.7 & 4.6
                 & 1.9 & \textbf{4.7} & 8.0 & 9.3 & 5.8
                 & 1.8 & {4.6} & 7.3 & 8.7 & 5.2 \\
       & MAPE & 23.0 & \textbf{21.5} & 18.5 & 11.7 & 14.0
                 & 27.4 & \textbf{25.5} & 21.9 & 13.9 & 16.4
                 & 25.8 & \textbf{23.9} & 20.0 & 13.2 & 15.4 \\
       & RMSE & 2.0 & \textbf{4.7} & 7.5 & 8.9 & 5.8
                 & 2.4 & \textbf{5.7} & 9.2 & 11.1 & 7.3
                 & 2.2 & \textbf{5.4} & 8.6 & 10.1 & 6.6 \\
\midrule
HYB(4) & MAE  & 1.8 & {3.9} & 6.9 & 7.8 & 5.3
                 & 2.1 & \textbf{4.7} & 8.6 & 9.4 & 6.5
                 & 2.0 & \textbf{4.5} & 7.8 & 8.9 & 5.9 \\
       & MAPE & 24.5 & 21.8 & 19.3 & 12.5 & 15.1
                 & 28.9 & 25.8 & 22.8 & 14.6 & 17.7
                 & 27.4 & 24.4 & 20.8 & 13.9 & 16.6 \\
       & RMSE & 2.2 & 4.8 & 8.1 & 9.0 & 6.6
                 & 2.6 & 5.9 & 9.9 & 11.3 & 8.1
                 & 2.5 & 5.6 & 9.1 & 10.3 & 7.2 \\
\midrule
HYB(5) & MAE  & 1.5 & 3.9 & 6.2 & 7.6 & 4.8
                 & 1.8 & 4.8 & 8.0 & 9.2 & 6.1
                 & 1.7 & 4.6 & 7.2 & 8.6 & 5.4 \\
       & MAPE & 22.8 & 22.0 & 17.9 & 12.0 & 13.7
                 & 26.9 & 26.9 & 21.2 & 14.3 & 16.1
                 & 24.6 & 25.4 & 19.9 & 13.6 & 15.0 \\
       & RMSE & 1.9 & 4.9 & 7.4 & 8.8 & 6.0
                 & 2.3 & 6.0 & 9.1 & 11.0 & 7.4
                 & 2.1 & 5.7 & 8.5 & 10.0 & 6.7 \\
\bottomrule
\end{tabular}
\end{table}

Table ~\ref{tab:hyb-sidebyside} presents a comparative evaluation of five distinct hybrid models, denote through HYB, across three bus routes and various Estimated Time of Arrival ranges (0–10, 10–25, 25–45, and 45+ minutes). Each model adheres to the HYB($n_m$) framework, wherein a meta-learner, implemented as a feed-forward neural network $(f_{NN})$, adaptively blends predictions from up to five base models. The hyperparameter $n_m$ critically regulates the number of active base models contributing to the final prediction. While HYB(1) occasionally registers the lowest absolute errors, a comprehensive analysis reveals that HYB(2) consistently emerges as the most effective configuration, demonstrating an optimal balance of accuracy, stability,computational complexity and theoretical congruence.

Key findings underscore the robust performance of HYB(2). It consistently achieves the lowest or near-lowest Mean Absolute Percentage Error across all evaluated routes (ranging from 12.3\% to 14.9\%), complemented by competitive Mean Absolute Error (MAE, 4.5–5.7 minutes) and Root Mean Square Error (RMSE, 5.4–6.9 minutes). This superior performance across metrics indicates a marked robustness compared to higher-$n_m$ models, which appear to suffer from increased noise or dilution of predictive signal. HYB(2) further demonstrates particular strength in short ETA ranges (0–10 minutes), where it records the lowest errors across all metrics, underscoring its suitability for critical real-time, high-frequency prediction scenarios. Crucially, its competitiveness extends to longer ETA ranges (25–45 and 45+ minutes), where it consistently maintains low errors despite the inherent increase in predictive uncertainty, affirming its versatility across temporal scales. Route-level insights reveal HYB(2)'s adaptability to diverse route complexities, showing particular excellence in Route 3. Even on Route 2, which generally exhibits the highest errors across all models, HYB(2) maintains it's stability of performane and computation complexity. This consistent efficacy is underpinned by its theoretical alignment: the $n_m=2$ configuration effectively balances model diversity and computational efficiency.
\begin{figure}[htbp]
    \centering
    \includegraphics[width=\linewidth]{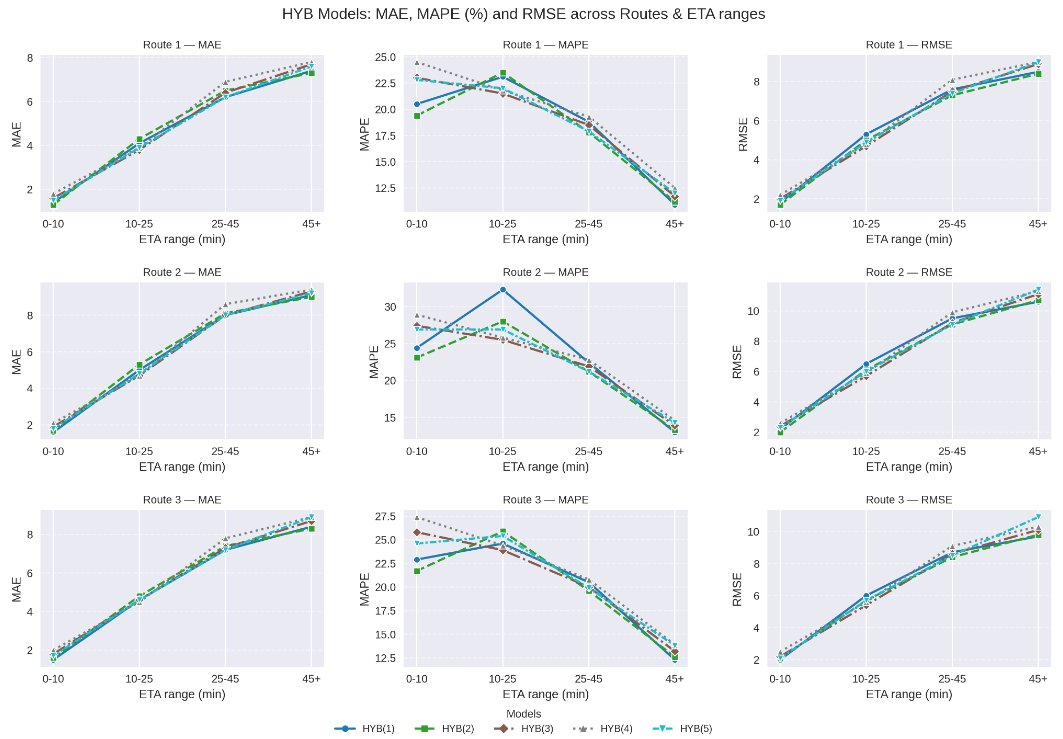}
    \caption{Comparative performance of five hybrid ETA prediction models (HYB(1)–HYB(5)) across three bus routes (rows) and three error metrics (columns). Each subplot depicts the variation of prediction errors across ETA ranges (0–10, 10–25, 25–45, and 45+ minutes). Metrics include Mean Absolute Error (MAE, minutes), Mean Absolute Percentage Error (MAPE, \%), and Root Mean Squared Error (RMSE, minutes). Distinct markers, colors, and line styles represent different hybrid configurations, where the meta-learner $f_{NN}$ integrates predictions from a varying number of base models.}
    \label{fig:hyb_downsize_analysis}
\end{figure}

\subsection{Comparison with Baseline Models}

Comparative analysis is used to evaluate proposed models against existing models specifically developed for graph time series forecasting.  To ensure a fair and consistent comparison, the current study investigated selected models that operate on fixed adjacency structures and can forecast one-dimensional graph signals.This restriction allows us to focus on methods that explicitly leverage the spatial dependencies 
encoded in a pre-defined graph, while modeling temporal dynamics for prediction. 
Formally, the model input is given as 
$\mathbb{R}^{N \times N \times 1 \times T}$, where $N$ denotes the number of nodes, 
$N \times N$ represents the fixed spatial graph adjacency, $1$ corresponds to the feature channel 
(In our case speed), and $T$ is the input time window length. 
The output is represented as $\mathbb{R}^{N \times N \times 1 \times U}$, where $U$ is the number of steps forecast into the future.
Some of the model overviews are given in table~\ref{tab:spatio_temporal_models}.

\begin{table}[h!]
\centering
\caption{Spatio-temporal GNN models with fixed adjacency: pros and cons.}
\label{tab:spatio_temporal_models}
\fontsize{5.9}{5.9}\selectfont
\renewcommand{\arraystretch}{1.90}
\setlength{\tabcolsep}{2.3pt}
\begin{tabular}{@{}p{4.2cm} p{4.3cm} p{4.3cm}@{}}
\toprule
\textbf{Model} & \textbf{Pros} & \textbf{Cons} \\ \midrule
DCRNN\cite{DBLP:journals/corr/LiYSL17} & Captures directed flows; principled diffusion modeling & RNN backbone; slow training, limited parallelism \\
STGCN\cite{Yu_2018} & Simple, efficient baseline & Weak long-range temporal modeling \\
GWNet (fixed A)\cite{wu2019graphwavenetdeepspatialtemporal} & Dilated convs capture long history; strong benchmark & Complex; tuning dilations is tricky \\
T-GCN\cite{Zhao_2020} & Simple GCN+GRU hybrid; effective on small data & RNN bottleneck; weak long-horizon \\
ASTGCN\cite{zhu2020astgcnattributeaugmentedspatiotemporalgraph} & Temporal+spatial attention; good for periodic patterns & Computationally heavy; hard to tune \\
STGAT (fixed A)\cite{9010834} & Flexible neighbor weighting & Overfits small datasets \\
GMAN (fixed A)\cite{zheng2019gmangraphmultiattentionnetwork} & Strong long-horizon accuracy; transformer-style & Very computationally expensive \\
MTGNN (fixed A)\cite{GAO202288} & Scalable, efficient mix-hop + dilated convs & Slightly weaker than adaptive version \\
ST-ResNet / DeepST\cite{zhang2017deepspatiotemporalresidualnetworks} & Great for city demand on grids & Poor for irregular graphs \\
STG2Seq\cite{bai2019stg2seqspatialtemporalgraphsequence} & Strong encoder–decoder; multi-step forecasting & Error accumulation in seq2seq \\
STFGNN\cite{li2021spatialtemporalfusiongraphneural} & Joint space-time fusion; balanced performance & Computationally heavy \\
Bi-GRCN\cite{Jiang2022Bi-GRCN} & Bi-directional capture; good short-term accuracy & Less robust for long-term horizons \\
STJGCN\cite{yan2018spatial} & Captures cross-time dependencies & Memory heavy for long horizons \\
USTGCN\cite{roy2021unifiedspatiotemporalmodelingtraffic} & Unified message passing; easier training & Underperforms attention-based models \\
ST-MetaNet (static)\cite{9096591} & Few-shot / transfer-friendly & Complex training procedure \\
PDFormer (static A)\cite{jiang2024pdformerpropagationdelayawaredynamic} & Transformer backbone; powerful for long horizons & Data hungry; high compute cost \\
STNorm\cite{10.1145/3447548.3467330} & Stabilizes training with normalisation & Incremental gains only \\
T-GCN++\cite{9786482} & Optimized for traffic datasets; easy deployment & Only modest improvements over T-GCN \\
ST-GConv (ChebNet)\cite{he2024convolutionalneuralnetworksgraphs} & Theoretically grounded spectral conv & Sensitive to Laplacian quality \\
Gated ST-GCN\cite{cheng2020gas} & Improves temporal stability with gating & Slightly higher complexity \\
ST Transformer (fixed A)\cite{Plizzari_2021} & Excellent for long horizons & Requires large data \\
MSGNN (fixed A)\cite{Qiu_2024} & Multi-scale spatial patterns & Higher memory and complexity \\
ST-GCN (residual+dilated)\cite{10662672} & Larger receptive field; strong baseline & Still static, lacks adaptivity \\
ST-Diffusion (fixed kernel)\cite{khan2025stdiffusion} & Good for asymmetric flows; principled & RNN-like, slower training \\
ConvLSTM+GCN\cite{GOSALA2025107657} & Captures spatial + temporal recurrence & Limited parallelism \\
Graph Residual FlowNet (fixed A)\cite{darvariu2024graphneuralmodelingnetwork} & Stable training; engineering-focused & Limited novelty \\
ST-MLP (fixed A)\cite{wang2023stmlpcascadedspatiotemporallinear} & Lightweight; surprisingly strong & Limited expressiveness \\ 
\bottomrule
\end{tabular}
\end{table}

Proposed framework HYB(2) was compared against a diverse set of established spatio-temporal graph models, namely DCRNN (classic diffusion convolution, strong benchmark, handles directed flows), STGCN (simple and efficient baseline, widely compared against), GWNet with fixed adjacency (dilated temporal convolutions, strong long-horizon baseline), T-GCN (GCN+GRU hybrid, effective for small/medium datasets), MTGNN with fixed adjacency (scalable mix-hop and dilated convolutions, efficient and practical), ST-ResNet / DeepST (grid-based model, strong for urban demand and flow prediction), STFGNN (fusion of spatial and temporal layers, balanced performance), and ST-GConv / ChebNet (spectral GCN variant, theoretically grounded). This set avoids attention/transformer-based approaches while still covering a broad spectrum: two classics (DCRNN, STGCN), two RNN-hybrids (T-GCN, GWNet), two scalable/practical models (MTGNN, STFGNN), one grid-based baseline (ST-ResNet /DeepST), and one spectral model (ST-GConv / ChebNet). Together, these eight models form a strong and representative benchmark suite to examine the performance of our framework.For input of the graph time series masked imputed speed at graph level is taken.
\[\mathcal{M}^{(t)} \odot \Omega_{\text{obs}}^{(t)} + (1 - \mathcal{M}^{(t)}) \odot \Omega_{\text{dft}}^{(t)}\]\

\begin{table}[h!]
\centering
\caption{Comparison of MAE, MAPE (\%), and RMSE for proposed model (HYB(2)) and baseline models across ETA ranges in Route 1, Route 2, and Route 3.(ETA computation is done for benchmark models using Algorithm~\ref{alg:benchmark_eta_computation}.)}
\label{tab:models-sidebyside}
\fontsize{4.9}{4.9}\selectfont
\renewcommand{\arraystretch}{1.20}
\setlength{\tabcolsep}{1.2pt}
\begin{tabular}{l l | @{\hskip 1pt}ccccc@{\hskip 1pt} | @{\hskip 1pt}ccccc@{\hskip 1pt} | @{\hskip 1pt}ccccc@{\hskip 1pt}}
\toprule
\multirow{2}{*}{\textbf{Model}} & \multirow{2}{*}{\textbf{Metric}}
& \multicolumn{5}{c|}{\textbf{Route 1}} 
& \multicolumn{5}{c|}{\textbf{Route 2}} 
& \multicolumn{5}{c}{\textbf{Route 3}} \\
\cmidrule(lr){3-7} \cmidrule(lr){8-12} \cmidrule(lr){13-17}
 & & 0--10 & 10--25 & 25--45 & 45+ & Overall
   & 0--10 & 10--25 & 25--45 & 45+ & Overall
   & 0--10 & 10--25 & 25--45 & 45+ & Overall \\
\midrule
\textbf{HYB(2) (Proposed)} & MAE  & 1.3 & 4.3 & 6.5 & \textbf{7.3} & 4.5
                 & 1.7 & 5.3 & \textbf{8.1} & \textbf{9.0} & 5.7
                 & \textbf{1.6} & 4.8 & 7.4 & \textbf{8.3} & \textbf{4.8} \\
       & MAPE & \textbf{19.4} & 23.5 & \textbf{17.8} & 11.2 & \textbf{12.3}
                 & \textbf{23.1} & 28.0 & \textbf{21.2} & 13.3 & \textbf{14.9}
                 & 21.7 & \textbf{25.9} & 19.6 & 12.6 & \textbf{13.8} \\
       & RMSE & 1.7 & 5.0 & \textbf{7.3} & \textbf{8.4} & \textbf{5.4}
                 & \textbf{2.0} & 6.1 & \textbf{9.1} & 10.7 & \textbf{6.9}
                 & 2.1 & 5.7 & \textbf{8.4} & 9.8 & 6.4 \\
\midrule
DCRNN & MAE  & \textbf{1.2} & \textbf{4.0} & 6.8 & 8.2 & 5.1
                 & \textbf{1.4} & 5.9 & 8.7 & 9.9 & 6.3
                 & 1.9 & 5.1 & 6.8 & 9.2 & 5.7 \\
       & MAPE & 22.1 & 22.6 & 19.9 & 12.1 & 14.1
                 & 25.2 & \textbf{27.2} & 22.4 & 14.2 & 16.4
                 & 21.1 & 26.8 & 21.4 & 12.0 & 16.8 \\
       & RMSE & \textbf{1.4} & \textbf{4.7} & 7.9 & 9.3 & 6.3
                 & 2.3 & 5.8 & 10.0 & 10.4 & 7.2
                 & 2.4 & 6.0 & 9.0 & 9.2 & 6.7 \\
\midrule
STGCN & MAE  & 2.2 & 4.9 & \textbf{6.2} & 8.8 & 5.4
                 & 2.3 & 6.2 & 9.3 & 10.5 & 6.9
                 & 2.2 & 5.7 & 8.0 & 10.1 & 6.3 \\
       & MAPE & 24.2 & 24.4 & 20.5 & 13.3 & 15.0
                 & 28.5 & 30.7 & 23.6 & 15.1 & 17.3
                 & 24.4 & 27.7 & 21.7 & 13.5 & 17.7 \\
       & RMSE & 2.6 & 5.6 & 8.2 & 9.9 & 6.9
                 & 2.9 & 7.4 & 11.0 & 11.9 & 8.3
                 & 2.7 & 6.6 & 9.5 & 10.9 & 7.4 \\
\midrule
GWNet (fixed A) & MAE  & 1.8 & 4.3 & 6.8 & 8.1 & 4.6
                 & 2.8 & \textbf{5.0} & 8.2 & 9.2 & \textbf{5.6}
                 & 2.3 & 4.9 & \textbf{7.1} & 8.7 & 5.0 \\
       & MAPE & 20.0 & 22.9 & 19.0 & \textbf{11.1} & 13.0
                 & 24.1 & 28.0 & 21.6 & 13.3 & 15.0
                 & \textbf{21.1} & 26.0 & \textbf{19.1} & \textbf{11.6} & 14.6 \\
       & RMSE & 1.9 & 5.0 & 7.6 & 8.8 & 5.6
                 & 2.2 & \textbf{5.7} & 9.1 & 10.8 & 7.1
                 & 2.1 & \textbf{5.6} & 8.5 & 9.9 & \textbf{6.1} \\
\midrule
T-GCN & MAE  & 2.6 & 4.6 & 6.7 & 7.8 & 4.9
                 & 2.9 & 5.7 & 8.4 & 9.4 & 6.2
                 & 1.8 & \textbf{4.6} & 7.7 & 8.8 & 5.4 \\
       & MAPE & 20.9 & 24.0 & 18.6 & 12.0 & 13.4
                 & 24.5 & 29.1 & 21.7 & 13.0 & 15.8
                 & 22.8 & 26.3 & 20.1 & 12.8 & 15.0 \\
       & RMSE & 2.0 & 5.3 & 7.7 & 8.9 & 6.0
                 & 2.3 & 6.5 & 9.5 & 10.0 & 7.4
                 & \textbf{1.9} & 6.1 & 8.7 & 10.1 & 6.8 \\
\midrule
MTGNN (fixed A) & MAE  & \textbf{1.3} & 4.5 & 6.6 & 7.4 & 4.6
                 & 1.7 & \textbf{5.2} & 8.2 & 9.1 & 5.7
                 & 2.6 & 4.8 & 7.5 & 8.6 & 5.0 \\
       & MAPE & 19.6 & 23.7 & 18.3 & 11.3 & 12.7
                 & 23.5 & 27.9 & 21.4 & \textbf{12.4} & 15.0
                 & 21.8 & 26.0 & 19.5 & 12.6 & 14.3 \\
       & RMSE & 1.8 & 5.1 & 7.5 & 8.6 & 5.5
                 & 2.0 & 6.2 & 9.2 & \textbf{9.7} & 7.0
                 & 2.1 & 5.8 & 8.5 & \textbf{9.0} & 6.5 \\
\midrule
STFGNN & MAE  & 1.5 & 4.6 & 6.7 & 7.6 & 4.7
                 & 2.8 & 5.4 & 8.3 & 9.3 & 5.9
                 & 1.7 & 4.9 & 7.6 & 8.7 & 5.2 \\
       & MAPE & 20.1 & 23.9 & 18.5 & 11.6 & 13.0
                 & 23.7 & 28.2 & 21.5 & 13.7 & 15.2
                 & 22.1 & 26.3 & 19.9 & 12.7 & 14.6 \\
       & RMSE & 1.9 & 5.2 & 7.6 & 8.7 & 5.7
                 & 2.1 & 6.2 & 9.3 & 10.8 & 7.1
                 & 2.2 & 5.9 & 8.6 & 9.9 & 6.6 \\
\midrule
ST-ResNet / DeepST & MAE  & 3.1 & 6.7 & 8.6 & 10.3 & 7.5
                 & 3.2 & 7.4 & 10.8 & 12.3 & 9.0
                 & 3.1 & 7.2 & 10.4 & 12.8 & 9.9 \\
       & MAPE & 33.8 & 33.1 & 25.6 & 17.8 & 21.9
                 & 36.6 & 37.3 & 29.9 & 20.5 & 22.4
                 & 34.0 & 36.1 & 26.8 & 18.3 & 21.3 \\
       & RMSE & 4.1 & 8.3 & 11.8 & 13.5 & 9.9
                 & 4.4 & 8.8 & 13.0 & 14.0 & 10.5
                 & 3.9 & 8.4 & 12.3 & 14.3 & 10.3 \\
\midrule
ST-GConv (ChebNet) & MAE  & 3.4 & 6.2 & 8.1 & 10.9 & 7.9
                 & 3.6 & 7.1 & 10.2 & 11.3 & 8.0
                 & 3.8 & 7.5 & 10.9 & 13.8 & 9.2 \\
       & MAPE & 32.0 & 31.6 & 24.4 & 16.9 & 20.7
                 & 34.2 & 35.5 & 28.1 & 19.3 & 21.2
                 & 31.9 & 34.0 & 25.6 & 19.4 & 20.4 \\
       & RMSE & 3.8 & 7.7 & 11.2 & 12.6 & 9.3
                 & 4.1 & 8.2 & 12.1 & 13.1 & 9.9
                 & 3.3 & 7.8 & 11.7 & 13.4 & 9.7 \\
\bottomrule
\end{tabular}
\end{table}
The results in Table~\ref{tab:models-sidebyside} exhibit diverse performance profiles, with each model's strengths and limitations corresponding to its architectural features, as detailed in the background.\\\\
Short-range predictions (0--10 and 10--25 minutes) are dominated by DCRNN, MTGNN, and HYB(2). DCRNN attains low MAE (e.g., Route 1, 0--10 min: 1.2; Route 2, 0--10 min: 1.4) and RMSE (e.g., Route 1, 0--10 min: 1.4; Route 1, 10--25 min: 4.7), capitalizing on its principled diffusion modeling for proximate temporal dependencies. MTGNN demonstrates comparable MAE (e.g., Route 1, 0--10 min: 1.3; Route 2, 0--10 min: 1.7) through scalable mix-hop convolutions, emphasizing efficiency and practicality. HYB(2) remains highly competitive, exhibiting robust MAE (e.g., Route 1, 0--10 min: 1.3; Route 3, 0--10 min: 1.6) and MAPE (e.g., Route 1, 0--10 min: 19.4), underscoring its hybrid architecture's efficacy in modeling short-term spatio-temporal dynamics.\\\\
Long-range predictions (25--45 and 45+ minutes) yield competitive outcomes among GWNet, HYB(2), and MTGNN. GWNet excels in MAPE (e.g., Route 1, 45+ min: 11.1; Route 3, 25--45 min: 19.1) via dilated convolutions for extended temporal histories, albeit constrained by tuning complexity. HYB(2) delivers strong MAE (e.g., Route 1, 25--45 min: 6.5; Route 2, 45+ min: 9.0) and RMSE (e.g., Route 2, 25--45 min: 9.1), whereas MTGNN competes effectively in RMSE (e.g., Route 2, 45+ min: 9.7; Route 3, 45+ min: 9.0). In contrast, T-GCN underperforms in long-range scenarios (e.g., Route 2, 25--45 min MAE: 8.4), attributable to its RNN bottleneck impeding extended-horizon forecasting.
Grid-based and spectral models (ST-ResNet and ST-GConv) consistently yield elevated errors (e.g., ST-ResNet MAE: 3.1--12.8; ST-GConv RMSE: 3.3--13.4) without leading in any category. These outcomes reflect inherent limitations: ST-ResNet's optimization for grid-structured urban demand renders it ill-suited for irregular graphs, while ST-GConv's performance is hampered by sensitivity to Laplacian quality.\\\\
HYB(2), MTGNN, and STFGNN exhibit balanced performance across ranges, with moderate errors in short- and long-term predictions. For instance, HYB(2) maintains consistent MAPE (e.g., Route 1, 25--45 min: 17.8), MTGNN in RMSE (e.g., Route 2, 0--10 min: 2.0), and STFGNN in MAE (e.g., Route 3, 0--10 min: 1.7). Their designs---HYB(2)'s hybridization, MTGNN's mix-hop convolutions, and STFGNN's spatio-temporal fusion---promote versatility, though STFGNN's computational demands and MTGNN's marginal underperformance relative to adaptive variants may constrain scalability.
MAPE underscores challenges in relative error mitigation, with all models displaying elevated values relative to MAE and RMSE, particularly ST-ResNet and ST-GConv (e.g., Route 2, 10--25 min: 37.3 and 35.5). GWNet, HYB(2), and MTGNN fare better in long-range MAPE (e.g., Route 1, 45+ min: 11.1--11.3), implying enhanced capacity for long-term relative error control through superior temporal pattern capture.\\\\
In summary, Table~\ref{tab:models-sidebyside} delineates a performance spectrum wherein DCRNN, MTGNN, and our HYB(2) excel in short-range forecasting, while GWNet, HYB(2), and MTGNN perform well in long-range predictions. HYB(2) is quite competitive with these best-performing models across all metrics and ranges, evincing robustness for both short- and long-term forecasting. ST-ResNet and ST-GConv prove less efficacious for irregular graphs, emphasizing the value of graph-aware architectures. These observations accord with the background's emphasis on fixed-adjacency models for one-dimensional graph signal forecasting, where variations stem from architectural merits (e.g., diffusion modeling, dilated convolutions) and drawbacks (e.g., computational complexity, RNN bottlenecks). Consequently, model selection should account for task-specific requirements, such as ETA range and graph topology, to maximize predictive accuracy and efficiency.
\begin{figure}[htbp]
    \centering
    \includegraphics[width=\linewidth]{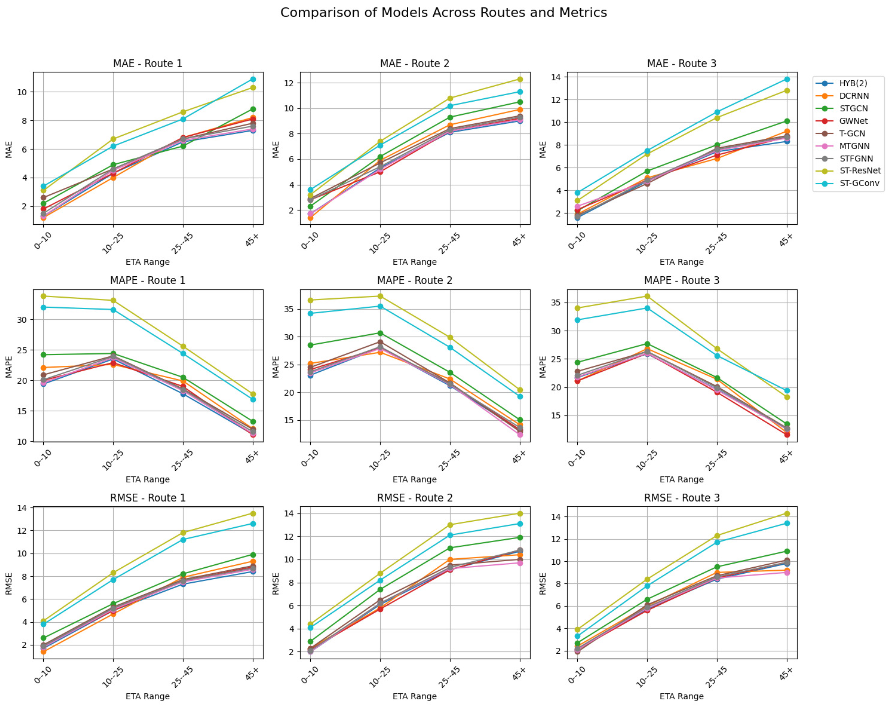}
    \caption{Comparative performance of HYB(2) (proposed) with eight models.
(Benchmark models display prediction patterns similar to HYB(2), as all fixed-adjacency methods rely on the same DFT-based initialization for imputing missing data, without which evaluation under incomplete observations is not feasible.)}
    \label{fig:indv_analysis}
\end{figure}

\begin{table}[h!]
\centering
\caption{Win Rate Analysis—Quantifying Model Dominance Across Prediction Scenarios. This table elucidates the comparative edge of each model by computing the win rate as the percentage of five predefined ETA prediction ranges where it registers the lowest error for specified metrics (MAE, MAPE, RMSE) on each of three bus routes (R1, R2, R3); values are derived through exhaustive pairwise comparisons within each range-metric-route triplet, with overall percentages aggregated as route-averaged means to highlight robust, context-specific leadership in bus arrival forecasting.}
\label{tab:win-rates-detailed}
\fontsize{4.9}{4.9}\selectfont
\renewcommand{\arraystretch}{1.20}
\setlength{\tabcolsep}{7.1pt}

\begin{tabular}{l|ccc|ccc|ccc|c}
\toprule
\textbf{Model:Win rate \%} 
& \multicolumn{3}{c|}{\textbf{MAE}} 
& \multicolumn{3}{c|}{\textbf{MAPE}} 
& \multicolumn{3}{c|}{\textbf{RMSE}} 
& \textbf{Overall (\%)} \\
\cmidrule(lr){2-4} \cmidrule(lr){5-7} \cmidrule(lr){8-10}
& \textbf{R1} & \textbf{R2} & \textbf{R3} 
& \textbf{R1} & \textbf{R2} & \textbf{R3} 
& \textbf{R1} & \textbf{R2} & \textbf{R3} 
& \\
\midrule

HYB(2) (Proposed)      & 40.0 & 40.0 & 60.0 & 60.0 & 60.0 & 40.0 & 60.0 & 60.0 & 20.0 & \textbf{48.9} \\
DCRNN                 & 40.0 & 20.0 & 20.0 & 20.0 & 20.0 & 20.0 & 40.0 & 0.0  & 0.0  & 20.0 \\
STGCN                 & 20.0 & 0.0  & 0.0  & 0.0  & 0.0  & 0.0  & 0.0  & 0.0  & 0.0  & 2.2  \\
GWNet (fixed A)       & 0.0  & 40.0 & 0.0  & 20.0 & 0.0  & 60.0 & 0.0  & 40.0 & 40.0 & 22.2 \\
T-GCN                 & 0.0  & 0.0  & 20.0 & 0.0  & 0.0  & 0.0  & 0.0  & 0.0  & 20.0 & 4.4  \\
MTGNN (fixed A)       & 20.0 & 0.0  & 0.0  & 0.0  & 20.0 & 0.0  & 0.0  & 40.0 & 20.0 & 11.1 \\
STFGNN                & 0.0  & 0.0  & 0.0  & 0.0  & 0.0  & 0.0  & 0.0  & 0.0  & 0.0  & 0.0  \\
ST-ResNet / DeepST    & 0.0  & 0.0  & 0.0  & 0.0  & 0.0  & 0.0  & 0.0  & 0.0  & 0.0  & 0.0  \\
ST-GConv (ChebNet)    & 0.0  & 0.0  & 0.0  & 0.0  & 0.0  & 0.0  & 0.0  & 0.0  & 0.0  & 0.0  \\

\bottomrule
\end{tabular}
\end{table}
Table \ref{tab:win-rates-detailed} summarizes model dominance using win rates across ETA ranges, routes, and error metrics. The proposed HYB(2) achieves the highest overall win rate (48.9\%), consistently outperforming all baselines across MAE, MAPE, and RMSE. 

Among competing methods, GWNet (fixed adjacency) and DCRNN show moderate but route-specific competitiveness, whereas STGCN, T-GCN, and MTGNN (fixed A) register only limited wins. Several spatial–temporal baselines fail to achieve dominance in any scenario. Overall, the results highlight that robust performance across diverse ETA horizons—rather than isolated accuracy gains—is essential,and demonstrate the effectiveness of HYB(2) across all prediction scenarios.

\subsection{Hyper Parameter Study}
\begin{table}[h!]
\centering
\caption{Optimal Hyperparameters and Ablation Insights Across Models. Only for models MGCN and MST-AV the performance for different hyperparameters are given in Appendix.~\ref{tab:mgcn-ablation},~\ref{tab:mst-av-performance}}
\label{tab:optimal_hyperparams}
\fontsize{5.9}{5.9}\selectfont
\renewcommand{\arraystretch}{1.20}
\setlength{\tabcolsep}{2.2pt}
\begin{tabular}{@{}p{1.0cm} p{1.4cm} p{1.4cm} p{4.7cm} p{3.7cm}@{}}
\toprule
\textbf{Model} & \textbf{Optimal Configuration} & \textbf{Performance} & \textbf{    Rationale} & \textbf{Not Studied} \\
\midrule
MST-AV & Grid size = 50 mtr$\times$50 mtr& MAE $\approx$ 8.0--8.4 min, Time $\approx$ 85.6--59.2 ms & Finer grids (50$\times$50) capture localised traffic variations in dense urban areas, improving path accuracy, but scaling to 100$\times$100 reduces MAE variance across routes by averaging over more historical traces per cell. This range avoids the computational explosion of sub-50 m bins while outperforming coarser ones (e.g., 1000$\times$1000) that lose route-specific details, prioritising low latency for battery-constrained apps. & Only average speed is taken for each grid. No other central tendency study (e.g., mode, median) is done. \\
\midrule
GDRN-DFT & $\Delta = 7$ days, $\delta = 15$ min, $\tau = 2$, $d = 16$ & MAE = 5.936 min, Time = 1.149 s & This configuration strikes an optimal balance by capturing weekly traffic cycles ($\Delta=7$ days) without excessive historical data overhead, while a moderate sampling interval ($\delta=15$ min) aligns with typical GPS update rates to reduce noise. The diffusion lag $\tau=2$ enables effective spatial smoothing of sparse signals, and $d=16$ provides sufficient Fourier detail for periodic patterns without amplifying high-frequency artifacts, minimising both error and computation.& $\epsilon_g = 1.5$ km/hr, $\tau_f=\tau_b$ taken as $\tau$ the same for forward and backward diffusion. \\
\midrule
KNN & $\delta = 5$ min, $d = 5$, $D = 12$ & MAE = 8.842 min, Time = 1.7 s & A finer sampling ($\delta=5$ min) preserves short-term dynamics in urban bus routes, where rapid changes occur, while $d=5$ embedding dimension captures essential nonlinearities in speed profiles without overparameterisation. Lifting to $D=12$ in the Koopman space linearises dynamics efficiently, avoiding the quadratic time scaling of higher $D$ values, thus optimizing for real-time inference on edge devices. & Number of intermediate layers for encoder = decoder = 2; $p$, $q$ (encoder) and $p'$, $q'$ (decoder) nodes optimized via Ray Tune (not manually studied); $\lambda_k = 1$ (regularization studied between prediction and reconstruction loss components). Activation function used: Swish.Swish provides the smooth differentiability required by Koopman theory while avoiding the gradient saturation of Tanh, enabling better propagation through the encoder-decoder architecture.\\
\midrule
FENN & $\delta = 10$ min, $b = 6$, $l = 3$, $k = 7$ & MAE = 8.271 min, Time = 0.791 s & The 10-min interval balances temporal granularity with feature stability, incorporating weather and location without aliasing. A compact sinusoidal basis ($b=6$) encodes diurnal cycles succinctly, reducing dimensionality, while 3 layers ($l=3$) provide depth for feature interactions without vanishing gradients. Sequence length $k=7$ aligns with hourly patterns, ensuring predictive power while keeping the model lightweight for mobile deployment. & Number of hidden layers studied in a shallow range, but number of nodes in hidden layers not studied manually; it was optimised via Ray Tune.Swish (Activation for hidden layers) offers a good balance between non-saturation and smooth gradients, enhancing feature interaction learning while maintaining computational efficiency.  \\
\midrule
MGCN & $\Delta T = 7$ days, $\delta_g = 10$ min, $L = 3$ & MAE = 7.39 min, Time = 1.039 s & Weekly history ($\Delta T=7$ days) encompasses recurring patterns like rush hours, enhancing imputation accuracy for masked graph nodes. Graph sampling at $\delta_g=10$ min maintains resolution for spatial dependencies, and 3 layers ($L=3$) propagate information across the MST-augmented graph sufficiently to leverage neighbors without over-smoothing, which occurs at $L=4$ and degrades performance on sparse bus networks. & No regularisation between L1 and L2, so $\lambda_m = 0$ (no smoothness enforced for DFT estimates). \\
\bottomrule
\end{tabular}
\end{table}
To ensure robust and efficient model performance, we conducted a comprehensive hyperparameter study across all individual methods. The primary objectives were twofold: (1) to identify optimal configurations that balance prediction accuracy (measured by Mean Absolute Error, MAE, in minutes) and computational efficiency (measured in seconds or milliseconds per inference); and (2) to analyze the sensitivity and interdependencies of key hyperparameters on these metrics, enabling informed trade-offs in deployment scenarios with varying computational constraints.For complete tables detailing all hyperparameters studied across their full ranges, see tables in Appendix B ~\ref{tab:mgcn-ablation},~\ref{tab:mst-av-performance}. The study employed a grid search approach over discrete hyperparameter ranges, selected based on domain knowledge and preliminary experiments to cover realistic operational scales (e.g., sampling intervals aligned with GPS update frequencies, history lengths reflecting short- to long-term traffic patterns). A complete summary of the optimal hyperparameter setup, the resulting performance, the rationale for why a configuration was chosen, and the parameters not studied is provided in Table ~\ref{tab:optimal_hyperparams}. Metrics were averaged across folds, with computation times recorded on a standardised hardware setup (NVIDIA RTX 6000 - Graphics RAM Size 48 GB) to ensure reproducibility.

All models demonstrated sensitivity to hyperparameter choices, particularly temporal resolution and architectural depth, with optimal configurations avoiding both underfitting and computational explosion. The study provides a rigorous foundation for selecting models based on specific deployment constraints and performance requirements.

\section{Conclusion}

The present study provides a comprehensive assessment of bus arrival time prediction models, resulting in the development of the HYB(2) hybrid system. Testing across three bus routes offers valuable insights for transit agencies. Individual models serve specific purposes: the MST-AV model, while less accurate, is very fast, easy to interpret, and runs efficiently on basic hardware, making it ideal for systems with limited resources. For high-precision needs, single models excel—KOOP-NET is best for short-term predictions, FENN handles sudden weather changes (but requires significant data), GDRN-DFT balances spatial and temporal accuracy reliably, and MGCN effectively combines real-time traffic with network structure. Yet, the HYB(2) framework acts as a smart fusion mechanism, combining these core strengths to achieve unmatched robustness and performance. The flexible HYB(k) design allows agencies to blend any number of base models to match their data, computing power, and specific needs, creating a custom balance between prediction power and practical cost. Further analysis confirms HYB(2) as the optimal choice, delivering top accuracy, stability, and efficiency across all tested conditions and routes. When compared to leading models, HYB(2) not only matches but surpasses those less suited to unpredictable bus networks, establishing itself as a top-tier solution. Ultimately, this work provides transit managers with a versatile toolkit: simple, cost-effective models like MST-AV or GDRN-DFT for straightforward needs, or the adaptable HYB(k) framework—with HYB(2) as its prime example—for those seeking the highest accuracy, resilience, and fit for their unique operational environment, paving the way for smarter and more reliable urban transit.

\section{Limitations and Future Work}
\label{sec:limitations-future-work}

Although the proposed HYB($k$) framework exhibits promising efficacy in bus arrival time estimation, key limitations temper its generalisability. The evaluation is done on only three Kolkata bus routes, potentially under-representing  infrastructure, and urban morphologies elsewhere; thus, broader validation across varied metropolitan contexts is essential. Additionally, reliance on GPS and weather data neglects exogenous factors like incidents, obstructions, or passenger loads, which may critically affect predictions. Technically, constituent models require refinements in preprocessing, loss weighting, regularisation, and latency mitigation.
Future work should advance context-adaptive hybridisation, dynamically selecting $k$ and modelling based on covariates like time or weather. Integrating multimodal data (e.g., real-time disruptions) with efficient ingestion protocols would enhance robustness. Model-specific improvements—via benchmarking, feature selection, and hyperparameter tuning—paired with edge-deployable meta-learners and caching, promise a scalable, equitable urban transit forecasting paradigm.

\phantomsection
\addcontentsline{toc}{section}{Acknowledgements}
\bmhead{Acknowledgements}
We thank the Transport Department, Government of West Bengal, for providing temporary access to their archived data sources for this research work.

\section*{Declarations}

\begin{itemize}
\item Funding : Not applicable
\item Conflict of interest/Competing interests: Authors have no conflict of interest.
\item Ethics approval and consent to participate : Not applicable
\item Consent for publication : Not applicable
\item Data availability : Gps data for 200 sessions for each route(1-1 Transformed):\url{https://github.com/pratham-payra/BATP-research-work/blob/master/all_routes_protected_new.csv}(Complete gps data is available on request.) , Bus stop data is made available in the link: \url{https://github.com/pratham-payra/BATP-research-work/blob/master/Route_informations}, Weather data is available via : \url{https://api.openweathermap.org/data/2.5/weather?lat={lat}&lon={lon}&appid={API key}}
\item Materials availability : Not applicable
\item Code availability : Code is available in the link : \url{https://github.com/pratham-payra/BATP-research-work/tree/master}
\item Author contribution : P.P. conceived the study and performed data analysis and code development. All authors contributed to result interpretation, manuscript review, and approved the final version.
\end{itemize}

\clearpage
\begin{appendices}

\section{Algorithms}\label{secA1}

The complete pseudocode of the proposed training and ETA computation procedures is provided below for reference.

\begin{algorithm}[H]
\caption{Data Acquisition and Preprocessing}
\label{alg:data_acq_preproc}
\scriptsize
\begin{algorithmic}[0]
\Require GPS data ($N$ days), bus stop dataset, OpenWeatherMap API key
\Ensure Preprocessed dataset: grid cells, graph $\mathcal{G}(\mathcal{V}, \mathcal{E})$, MST, adjacency matrix $A$, weather $\mathcal{W}$

\State \textbf{Data Acquisition:}
\State $\triangleright$ Fetch GPS traces: timestamps, coordinates $(\vartheta, \varphi)$, Session ID, route, Unix timestamp
\State $\triangleright$ Fetch bus stop data: geo-referenced stops (level, name, route tags)
\For{each GPS timestamp $t$}
    \State $\triangleright$ Fetch weather via API: \url{https://api.openweathermap.org/data/2.5/weather}, store $\mathcal{W}_t = (\varrho_t (\%\%), \kappa_t (\text{mm}), \varpi_t (\text{K}), \mu_t (\%\%), \beta_t (\text{hPa}), \varpi^a_t (\text{K}), \nu_t (\text{m/s}))$
\EndFor

\State \textbf{Data Preprocessing:}
\For{each consecutive GPS pair $(\vartheta_1, \varphi_1)$, $(\vartheta_2, \varphi_2)$}
    \State $\triangleright$ Compute Haversine distance $d$: $a = \sin^2\left(\frac{\vartheta_2 - \vartheta_1}{2}\right) + \cos(\vartheta_1)\cos(\vartheta_2)\sin^2\left(\frac{\varphi_2 - \varphi_1}{2}\right)$, $c = 2 \arctan2(\sqrt{a}, \sqrt{1-a})$, $d = 6371 \cdot c$ (km)
    \State $\triangleright$ Compute speed $v = d / \Delta t$ ($\Delta t$ in hours)
\EndFor
\State $\triangleright$ Build $50 \text{m} \times 50 \text{m}$ grid: round $(\vartheta, \varphi)$ to $\vartheta_{\text{rounded}} = \text{round}(\vartheta / 0.0005) \cdot 0.0005$, $\varphi_{\text{rounded}} = \text{round}(\varphi / 0.0005) \cdot 0.0005$
\State $\triangleright$ Group GPS points by grid, compute median $(\vartheta^{\text{median}}, \varphi^{\text{median}})$
\State $\triangleright$ Form graph $\mathcal{G}(\mathcal{V}, \mathcal{E})$: vertices as cells, edges by Haversine distances
\State $\triangleright$ Extract MST (Kruskal's/Prim's algorithm)
\State $\triangleright$ Compute shortest path $\mathcal{P}$ from $\varsigma_{\text{start}}$ to $\varsigma_{\text{end}}$ (Dijkstra's algorithm, minimize $\sum_{e \in \mathcal{P}} d_e$)
\State $\triangleright$ Encode $A$: $A_{ij} = d_{ij}$ if $(i,j)$ in MST/path, else $0$
\State $\triangleright$ Form weather vectors $\mathcal{W}(i)_t = (\varrho(i)_t, \kappa(i)_t, \varpi_t, \mu_t, \beta_t, \varpi^a_t, \nu_t) \in \mathbb{R}^7$ for each $t, i$
\State $\triangleright$ Validate units/constraints (e.g., $\varpi_t > 0$, $\mu_t \in [0, 100]\%\%$)
\State $\triangleright$ Return dataset: GPS (speeds, grid), $\mathcal{G}$, MST, $A$, $\mathcal{W}$
\end{algorithmic}
\end{algorithm}

\begin{algorithm}[H]
\caption{Training Algorithm for MST-AV: Compute Historical Mean Speeds}
\label{alg:mst_av_training}
\scriptsize
\begin{algorithmic}[0]
\Require Historical GPS data with speeds $v$ for each node $i \in \mathcal{V}$ over multiple traces
\Ensure Historical mean speeds $\mathcal{\widehat{U}}_i^{\text{avg}}$ for all nodes $i \in \mathcal{V}$
\State $\rightarrow$ Initialize sum and count for each node: $\text{sum}_i \gets 0$, $\text{count}_i \gets 0$ for all $i \in \mathcal{V}$
\For{each historical trace}
    \For{each node $i$ visited in the trace with speed $v_i$}
        \State $\rightarrow$ Update $\text{sum}_i \gets \text{sum}_i + v_i$
        \State $\rightarrow$ Update $\text{count}_i \gets \text{count}_i + 1$
    \EndFor
\EndFor
\For{each node $i \in \mathcal{V}$}
    \If{$\text{count}_i > 0$}
        \State $\rightarrow$ Compute $\mathcal{\widehat{U}}_i^{\text{avg}} \gets \text{sum}_i / \text{count}_i$
    \Else
        \State $\rightarrow$ Set $\mathcal{\widehat{U}}_i^{\text{avg}} \gets$ default value (e.g., global average or 0)
    \EndIf
\EndFor
\State $\rightarrow$ Return mean speeds $\mathcal{\widehat{U}}_i^{\text{avg}}$ for all $i$
\end{algorithmic}
\end{algorithm}

\begin{algorithm}[H]
\caption{Computation Algorithm for MST-AV: Bus Arrival Time Estimation}
\label{alg:mst_av_computation}
\scriptsize
\begin{algorithmic}[0]
\Require Bus stop $(\vartheta_{\text{bs}}, \varphi_{\text{bs}})$, current bus $(\vartheta_{\text{bl}}, \varphi_{\text{bl}})$, graph $\mathcal{G}=(\mathcal{V},\mathcal{E})$, mean speeds $\mathcal{\widehat{U}}_i^{\text{avg}}$
\Ensure Estimated Time of Arrival ($\mathcal{ETA}_{MST-AV}$)
\State $\rightarrow$ Identify grid cell $G_{\text{bl}} \in \mathcal{V}$ for $(\vartheta_{\text{bl}}, \varphi_{\text{bl}})$
\State $\rightarrow$ Identify grid cell $G_{\text{bs}} \in \mathcal{V}$ for $(\vartheta_{\text{bs}}, \varphi_{\text{bs}})$
\State $\rightarrow$ Compute MST-augmented graph from $\mathcal{G}$ to get path set $\mathcal{P}$
\State $\rightarrow$ Extract shortest path $\mathcal{P}' = (G_{\text{bl}} \rightsquigarrow G_{\text{bs}}) \subseteq \mathcal{P}$ using Haversine-weighted edges
\State $\rightarrow$ Initialize $\mathcal{ETA}_{MST-AV} \gets 0$
\For{each edge $(i,j) \in \mathcal{P}'$}
    \State $\rightarrow$ Compute edge length $\mathcal{EL}_{ij} \gets \mathcal{H}\mathit{aversine}((\vartheta_i^{\text{median}}, \varphi_i^{\text{median}}), (\vartheta_j^{\text{median}}, \varphi_j^{\text{median}}))$
    \State $\rightarrow$ Compute average velocity $\mathcal{\widehat{U}}_{ij}^{\text{avg}} \gets (\mathcal{\widehat{U}}_i^{\text{avg}} + \mathcal{\widehat{U}}_j^{\text{avg}}) / 2$
    \State $\rightarrow$ Update $\mathcal{ETA}_{MST-AV} \gets \mathcal{ETA}_{MST-AV} + \mathcal{EL}_{ij} / \mathcal{\widehat{U}}_{ij}^{\text{avg}}$
\EndFor
\State $\rightarrow$ Return $\mathcal{ETA}_{MST-AV}$
\end{algorithmic}
\end{algorithm}

\begin{algorithm}[H]
\caption{Training Algorithm for GDRN-DFT: Impute and Analyze Speeds}
\label{alg:gdrn_dft_training}
\scriptsize
\begin{algorithmic}[0]
\Require Historical speeds $\mathbf{\Omega}^{(t_s)} \in \mathbb{R}^n$ for $t_s=1,\dots,T$ ($\Delta_g$(Time duration to train) = $T_{ref}-T_0$,$\delta_g$(time step duration),$T=\Delta_g/\delta_g$), $\epsilon_g$(cut-off to ignore low amplitude patterns),mask $\mathbf{\mathcal{M}}^{(t_s)} \in \{0,1\}^n$, graph adjacency $\mathbf{A}$
\Ensure DFT parameters: amplitudes $\mathcal{A}_r(k)$, phases $\phi_r^{(k)}$ for each node $r$
\State $\rightarrow$ Compute degree matrix $\mathbf{D}$, normalized Laplacian $\mathbf{L} = \mathbf{I} - \mathbf{D}^{-1/2} \mathbf{A} \mathbf{D}^{-1/2}$
\State $\rightarrow$ Initialize GDRN parameters: bi-LSTM weights $\mathbf{W}_*$, biases $\mathbf{b}_*$, output $\mathbf{W}_h, \mathbf{b}_h$
\For{each training epoch ($1\rightarrow d_g$)}
    \For{each time step $t_s$}
        \State $\rightarrow$ Compute diffused signal $\tilde{\mathbf{\Omega}}^{(t_s)} = \left( \frac{e^{-\tau_f \mathbf{L}} + e^{-\tau_b \mathbf{L}}}{2} \right) \mathbf{\Omega}^{(t_s)}$ (approx. via Chebyshev/Taylor)
        \State $\rightarrow$ Mask and impute: $\tilde{\mathbf{\Omega}}^{(t_s)} = \mathbf{\mathcal{M}}^{(t_s)} \odot \mathbf{\Omega}^{(t_s)} + (1 - \mathbf{\mathcal{M}}^{(t_s)}) \odot$ diffused
        \State $\rightarrow$ Forward LSTM: $\overrightarrow{\mathbf{h}}_{t_s} = \text{LSTM}_{\text{forward}}(\tilde{\mathbf{\Omega}}^{(t_s)}, \overrightarrow{\mathbf{h}}_{t_s-1})$
        \State $\rightarrow$ Backward LSTM: $\overleftarrow{\mathbf{h}}_{t_s} = \text{LSTM}_{\text{backward}}(\tilde{\mathbf{\Omega}}^{(t_s)}, \overleftarrow{\mathbf{h}}_{t_s+1})$
        \State $\rightarrow$ Concat $\mathbf{h}_{t_s} = [\overrightarrow{\mathbf{h}}_{t_s}; \overleftarrow{\mathbf{h}}_{t_s}]$
        \State $\rightarrow$ Predict $\hat{\mathbf{\Omega}}^{(t_s)} = \mathbf{W}_h \mathbf{h}_{t_s} + \mathbf{b}_h$
        \State $\rightarrow$ Compute loss $\mathcal{L}_{gd} = \left\| \mathbf{\mathcal{M}}^{(t_s)} \odot (\mathbf{\Omega}^{(t_s)} - \hat{\mathbf{\Omega}}^{(t_s)}) \right\|_2^2$
        \State $\rightarrow$ Update parameters via backpropagation
    \EndFor
\EndFor
\State $\rightarrow$ Impute full dataset: for each $t_s$, $\mathbf{\Omega}^{(t_s)} = \mathbf{\mathcal{M}}^{(t_s)} \odot \mathbf{\Omega}^{(t_s)} + (1 - \mathbf{\mathcal{M}}^{(t_s)}) \odot \hat{\mathbf{\Omega}}^{(t_s)}$
\For{each node $r$}
    \State $\rightarrow$ Extract time series $\rho_r = \{v_r^{(0)}, \dots, v_r^{(T-1)}\}$
    \State $\rightarrow$ Compute DFT: $\mathcal{C}_r(k) = \sum_{n=0}^{T-1} v_r^{(n)} e^{-i \frac{2\pi}{T} kn}$ for $k=0,\dots,T-1$
    \State $\rightarrow$ Compute amplitude $\mathcal{A}_r(k) = \frac{|\mathcal{C}_r(k)|}{T}$, 
phase $\phi_r^{(k)} = \tan^{-1}\!\left(\frac{\Im(\mathcal{C}_r(k))}{\Re(\mathcal{C}_r(k))}\right)$ 
\State \hspace{1.5em} If $\mathcal{A}_r(k) < \epsilon_g$ then set $\mathcal{A}_r(k) = 0$ else keep $\mathcal{A}_r(k)$
    
\EndFor
\State $\rightarrow$ Return DFT parameters $\mathcal{A}_r(k)$, $\phi_r^{(k)}$ for all $r, k$
\end{algorithmic}
\end{algorithm}

\begin{algorithm}[H]
\caption{Computation Algorithm for GDRN-DFT: Bus Arrival Time Estimation}
\label{alg:gdrn_dft_computation}
\scriptsize
\begin{algorithmic}[0]
\Require Bus stop $(\vartheta_{\text{bs}}, \varphi_{\text{bs}})$, current bus $(\vartheta_{\text{bl}}, \varphi_{\text{bl}})$, graph $\mathcal{G}=(\mathcal{V},\mathcal{E})$, DFT parameters $\mathcal{A}_i(k)$, $\phi_i^{(k)}$, time params $T, \delta, T_0, V_{\max}, \epsilon$
\Ensure Estimated Time of Arrival ($\mathcal{ETA}_{GDRN-DFT}$)
\State $\rightarrow$ Identify grid cell $G_{\text{bl}} \in \mathcal{V}$ for $(\vartheta_{\text{bl}}, \varphi_{\text{bl}})$
\State $\rightarrow$ Identify grid cell $G_{\text{bs}} \in \mathcal{V}$ for $(\vartheta_{\text{bs}}, \varphi_{\text{bs}})$
\State $\rightarrow$ Compute MST-augmented graph from $\mathcal{G}$ to get path set $\mathcal{P}$
\State $\rightarrow$ Extract shortest path $\mathcal{P}' = (G_{\text{bl}} \rightsquigarrow G_{\text{bs}}) \subseteq \mathcal{P}$ using Haversine-weighted edges
\State $\rightarrow$ Initialize $\mathcal{ETA}_{GDRN-DFT} \gets 0$, $t \gets$ current time
\For{each edge $(i,j) \in \mathcal{P}'$}
    \State $\rightarrow$ Compute $t_s \gets (t - T_0) / \delta$
    \State $\rightarrow$ Reconstruct speeds: $\mathcal{U}_i(t_s) = \sum_{k=0}^{T-1} \mathcal{A}_i(k) \cos(2\pi k t_s / T + \phi_i^{(k)})$
    \State $\rightarrow$ $\mathcal{U}_j(t_s) = \sum_{k=0}^{T-1} \mathcal{A}_j(k) \cos(2\pi k t_s / T + \phi_j^{(k)})$
    \State $\rightarrow$ Filter: $\widehat{\mathcal{U}}_i(t_s) = \max(\epsilon, \min(V_{\max}, \mathcal{U}_i(t_s)))$, similarly for $j$
    \State $\rightarrow$ Average velocity $\widehat{\mathcal{U}}_{ij}(t_s) = (\widehat{\mathcal{U}}_i(t_s) + \widehat{\mathcal{U}}_j(t_s)) / 2$
    \State $\rightarrow$ Compute edge length $\mathcal{EL}_{ij} \gets \mathcal{H}\mathit{aversine}((\vartheta_i^{\text{median}}, \varphi_i^{\text{median}}), (\vartheta_j^{\text{median}}, \varphi_j^{\text{median}}))$
    \State $\rightarrow$ Update $\mathcal{ETA}_{GDRN-DFT} \gets \mathcal{ETA}_{GDRN-DFT} + \mathcal{EL}_{ij} / \widehat{\mathcal{U}}_{ij}(t_s)$
    \State $\rightarrow$ Update $t \gets t + \mathcal{EL}_{ij} / \widehat{\mathcal{U}}_{ij}(t_s)$
\EndFor
\State $\rightarrow$ Return $\mathcal{ETA}_{GDRN-DFT}$
\end{algorithmic}
\end{algorithm}

\begin{algorithm}[H]
\caption{Training Algorithm for KNN: Train Koopman Neural Network}
\label{alg:knn_training}
\scriptsize
\begin{algorithmic}[0]
\Require Historical average speeds $\{v_m\}$ over time, window size $k(k=d_k)$, max speed $V_{\max}$, $\epsilon$, hyperparameters $p (auto-via-RayTune), q (auto-via-RayTune),p' (auto-via-RayTune), q' (auto-via-RayTune), D_k(dimension-of-lifted-space),\lambda_k(regularization-between-prediction-and-reconstruction),\delta_k(time-step-length)$
\Ensure Trained KNN parameters: encoder $\psi_\theta$, Koopman $\mathcal{K}_\phi$, decoder $\psi_{\theta'}^{-1}$
\State $\rightarrow$ Aggregate speeds into time series at intervals $\delta_k$
\For{each training sample}
    \State $\rightarrow$ Form input sequence $\mathbf{\Omega}_t = [v_{t-(k-1)}, \dots, v_t]$,
 Filter : $v_j' = min(V_{max}-\epsilon,max(v_j,\epsilon))$   \State $\rightarrow$ Logit transform: $\omega_j = \ln(v'_j / (V_{\max} - v'_j))$
    \State $\rightarrow$ Form $\mathbf{x}_t = [\omega_{t-(k-1)}, \dots, \omega_t]$
    \State $\rightarrow$ Form target $\mathbf{x}_{t+\Delta} = [\omega_{t+1}, \dots, \omega_{t+k}]$
\EndFor
\State $\rightarrow$ Initialize encoder MLP(Swish activation) $\psi_\theta$ (layers: input $\to p \to q \to D_k$)
\State $\rightarrow$ Initialize Koopman matrix $\mathcal{K}_\phi \in \mathbb{R}^{D_k \times D_k}$
\State $\rightarrow$ Initialize decoder MLP(Swish activation) $\psi_{\theta'}^{-1}$ (layers: $D_k \to q' \to p' \to$ output)
\For{each training epoch}
    \For{each batch $\mathbf{x}_t$}
        \State $\rightarrow$ Lift: $h_t = \psi_\theta(\mathbf{x}_t)$
        \State $\rightarrow$ Reconstruct: $\hat{\mathbf{x}}_t = \psi_{\theta'}^{-1}(h_t)$
        \State $\rightarrow$ Reconstruction loss: $\|\mathbf{x}_t - \hat{\mathbf{x}}_t\|_2^2$
        \State $\rightarrow$ For each step $\Delta$: Apply dynamics $h_{t+\Delta} = \mathcal{K}_\phi^\Delta h_t$
        \State $\rightarrow$ Predict: $\hat{\mathbf{x}}_{t+\Delta} = \psi_{\theta'}^{-1}(h_{t+\Delta})$
        \State $\rightarrow$ Prediction loss: $\sum_\Delta \|\mathbf{x}_{t+\Delta} - \hat{\mathbf{x}}_{t+\Delta}\|_2^2$
        \State $\rightarrow$ Total loss: $\mathcal{L}_{\text{KNN}} = $ reconstruction + $\lambda_k \times$ prediction
        \State $\rightarrow$ Update parameters via backpropagation
    \EndFor
\EndFor
\State $\rightarrow$ Return trained KNN model
\end{algorithmic}
\end{algorithm}

\begin{algorithm}[H]
\caption{Computation Algorithm for KNN: Bus Arrival Time Estimation}
\label{alg:knn_computation}
\scriptsize
\begin{algorithmic}[0]
\Require Bus stop $(\vartheta_{\text{bs}}, \varphi_{\text{bs}})$, current bus $(\vartheta_{\text{bl}}, \varphi_{\text{bl}})$, graph $\mathcal{G}=(\mathcal{V},\mathcal{E})$, trained KNN, $k(k=d_k), \delta_k, V_{\max}, \epsilon$
\Ensure Estimated Time of Arrival ($\mathcal{ETA}_{KNN}$)
\State $\rightarrow$ Identify $G_{\text{bl}} \in \mathcal{V}$ for $(\vartheta_{\text{bl}}, \varphi_{\text{bl}})$
\State $\rightarrow$ Identify $G_{\text{bs}} \in \mathcal{V}$ for $(\vartheta_{\text{bs}}, \varphi_{\text{bs}})$
\State $\rightarrow$ Compute MST-augmented graph from $\mathcal{G}$ to get $\mathcal{P}$
\State $\rightarrow$ Extract shortest path $\mathcal{P}' = (G_{\text{bl}} \rightsquigarrow G_{\text{bs}}) \subseteq \mathcal{P}$
\State $\rightarrow$ Compute total distance $\mathcal{RD} = \sum_{(i,j) \in \mathcal{P}'} \mathcal{EL}_{ij}$, where $\mathcal{EL}_{ij} = \mathcal{H}\mathit{aversine}((\vartheta_i^{\text{median}}, \varphi_i^{\text{median}}), (\vartheta_j^{\text{median}}, \varphi_j^{\text{median}}))$
\State $\rightarrow$ Initialize $\mathcal{ETA}_{KNN} \gets 0$, $t \gets$ current time step
\State $\rightarrow$ Set initial input speeds $[v_{t-(k-1)}, \dots, v_t]$
\While{$\mathcal{RD} > \epsilon$}
    \State $\rightarrow$ Filter: $v'_j = \min(V_{\max} - \epsilon, \max(v_j, \epsilon))$
    \State $\rightarrow$ Logit: $\omega_j = \ln(v'_j / (V_{\max} - v'_j))$
    \State $\rightarrow$ Form $\mathbf{x}_t = [\omega_{t-(k-1)}, \dots, \omega_t]$
    \State $\rightarrow$ Predict $\hat{\mathbf{x}}_{t+\Delta} = \psi_{\theta'}^{-1} (\mathcal{K}_\phi^\Delta \psi_\theta (\mathbf{x}_t))$
    \State $\rightarrow$ Inverse logit: $v_j = V_{\max} / (1 + e^{-\omega_j})$ for $\hat{\mathbf{x}}_{t+\Delta}$ to get $[v_{t+1}, \dots, v_{t+k}]$
    \For{each $v_j$ in $[v_{t+1}, \dots, v_{t+k}]$}
        \State $\rightarrow$ Update $\mathcal{RD} \gets \mathcal{RD} - v_j \cdot \delta$
        \State $\rightarrow$ Update $\mathcal{ETA}_{KNN} \gets \mathcal{ETA}_{KNN} + \delta$
        \If{$\mathcal{RD} < \epsilon$}
            \State $\rightarrow$ Break
        \EndIf
    \EndFor
    \State $\rightarrow$ Update input to predicted speeds, $t \gets t + k$
\EndWhile
\State $\rightarrow$ Return $\mathcal{ETA}_{KNN}$
\end{algorithmic}
\end{algorithm}
\begin{algorithm}[H]
\caption{Training Algorithm for FENN: Train Feature-Encoded Neural Network}
\label{alg:fenn_training}
\scriptsize
\begin{algorithmic}[0]
\Require Historical data: locations $(\vartheta_i, \varphi_i)$, weather $\mathcal{W}_i$, times $t_i$, speeds $v_i$, window $k$, $V_{\max}$, hyperparameters : $l_f(number-of-layers-for-Neural-Network), b(sin-feature-dim-of-time),\delta_n(time-step-length)$
\Ensure Trained FENN: encoder, neural network $\mathcal{F}_{NN}$
\For{each timestamp $t_i = t_0 + i \cdot \delta_n$ , $i:1\rightarrow k(k=(t-t_0)/\delta_n)$}
    \State $\rightarrow$ Fetch weather $\mathcal{W}_i$ via API for $(\vartheta_i, \varphi_i)$
    \State $\rightarrow$ Compute sines: $\varsigma_i(m) = \sin(\alpha_i + 2m\pi / b)$ for $m=0,\dots,b-1$, $\alpha_i = 2\pi / 24 \cdot \text{hour}(t_i)$
    \State $\rightarrow$ Aggregate speed $v_i$
    \State $\rightarrow$ Form feature $\mathbf{y_i} = (\vartheta_i, \varphi_i, \mathcal{W}_i, \varsigma_i(0), \dots, \varsigma_i(b-1), v_i) \in \mathbb{R}^{\gamma \times 1}$ 
\EndFor
\For{each training sample}
    \State $\rightarrow$ Form input matrix $\mathbf{Y_t} = [\mathbf{y_{t-(k-1)}}, \dots, \mathbf{y_t}] \in \mathbb{R}^{\gamma \times k}$
    \State $\rightarrow$ Encode: $\zeta_i = \text{encoder}(\mathbf{y_i})$ for each column, form $\mathbf{Z_t} = [\zeta_{t-(k-1)}, \dots, \zeta_t]$
    \State $\rightarrow$  Filter: $v'_j = \min(V_{\max} - \epsilon, \max(v_j, \epsilon))$ , Logit transform targets : $\omega_j = \ln(v_j' / (V_{\max} - v_j'))$ for $v'_{t+1},\dots,v'_{t+k}$, form $\mathbf{X_{t+\Delta}} = [\omega_{t+1}, \dots, \omega_{t+k}]$
\EndFor
\State $\rightarrow$ Initialize encoder (to scalar), NN with layers $k \to NN(Number-of-hidden-layers=l_f) \to k$
\For{each training epoch}
    \For{each batch $\mathbf{Z_t}$}
        \State $\rightarrow$ Predict $\hat{\mathbf{X}}_{t+\Delta} = \mathcal{F}_{NN}(\mathbf{Z_t})$ $Activation-function:Swish$
        \State $\rightarrow$ Compute loss $\mathcal{L}_{\text{FN}} = \sum_{\Delta} \|\mathbf{X_{t+\Delta}} - \hat{\mathbf{X}}_{t+\Delta}\|_2^2$
        \State $\rightarrow$ Update parameters via backpropagation
    \EndFor
\EndFor
\State $\rightarrow$ Return trained FENN model
\end{algorithmic}
\end{algorithm}

\begin{algorithm}[H]
\caption{Computation Algorithm for FENN: Bus Arrival Time Estimation}
\label{alg:fenn_computation}
\scriptsize
\begin{algorithmic}[0]
\Require Bus stop $(\vartheta_{\text{bs}}, \varphi_{\text{bs}})$, current bus $(\vartheta_{\text{bl}}, \varphi_{\text{bl}})$, graph $\mathcal{G}=(\mathcal{V},\mathcal{E})$, trained FENN, $k, \delta_n, V_{\max}, \epsilon, b$
\Ensure Estimated Time of Arrival ($\mathcal{ETA}_{FE-NN}$)
\State $\rightarrow$ Identify $G_{\text{bl}} \in \mathcal{V}$ for $(\vartheta_{\text{bl}}, \varphi_{\text{bl}})$
\State $\rightarrow$ Identify $G_{\text{bs}} \in \mathcal{V}$ for $(\vartheta_{\text{bs}}, \varphi_{\text{bs}})$
\State $\rightarrow$ Compute MST-augmented graph from $\mathcal{G}$ to get $\mathcal{P}$
\State $\rightarrow$ Extract shortest path $\mathcal{P}' = (G_{\text{bl}} \rightsquigarrow G_{\text{bs}}) \subseteq \mathcal{P}$
\State $\rightarrow$ Compute total distance $\mathcal{RD} = \sum_{(i,j) \in \mathcal{P}'} \mathcal{EL}_{ij}$, where $\mathcal{EL}_{ij} = \mathcal{H}\mathit{aversine}((\vartheta_i^{\text{median}}, \varphi_i^{\text{median}}), (\vartheta_j^{\text{median}}, \varphi_j^{\text{median}}))$
\State $\rightarrow$ Initialize $\mathcal{ETA}_{FE-NN} \gets 0$, $t \gets$ current time step
\State $\rightarrow$ Set initial input $\mathbf{Y_t} = [\mathbf{y_{t-(k-1)}}, \dots, \mathbf{y_t}]$ (with current location, weather, sines, speeds)
\While{$\mathcal{RD} > \epsilon$}
    \State $\rightarrow$ Encode $\mathbf{Z_t} = [\text{encoder}(\mathbf{y_{t-(k-1)}}), \dots, \text{encoder}(\mathbf{y_t})]$
    \State $\rightarrow$ Predict $\hat{\mathbf{X}}_{t+\Delta} = \mathcal{F}_{NN}(\mathbf{Z_t})$
    \State $\rightarrow$ Inverse logit: $v_j = V_{\max} / (1 + e^{-\omega_j})$ for $\hat{\mathbf{X}}_{t+\Delta}$ to get $[v_{t+1}, \dots, v_{t+k}]$
    \For{each $v_j$ in $[v_{t+1}, \dots, v_{t+k}]$}
        \State $\rightarrow$ Update $\mathcal{RD} \gets \mathcal{RD} - v_j \cdot \delta_n$
        \State $\rightarrow$ Update $\mathcal{ETA}_{FE-NN} \gets \mathcal{ETA}_{FE-NN} + \delta_n$
        \If{$\mathcal{RD} < \epsilon$}
            \State $\rightarrow$ Break
        \EndIf
    \EndFor
    \State $\rightarrow$ Update $\mathbf{Y_{t+k}}$: fetch new $\mathcal{W}_i$ via API for updated locations along $\mathcal{P}'$ using $v_j \cdot \delta_n$, update $(\vartheta_i, \varphi_i)$, sines with $t_i + j \cdot \delta_n$, speeds with $v_{t+j}$
    \State $\rightarrow$ Set $\mathbf{Y_t} \gets \mathbf{Y_{t+k}}$, $t \gets t + k$
\EndWhile
\State $\rightarrow$ Return $\mathcal{ETA}_{FE-NN}$
\end{algorithmic}
\end{algorithm}

\begin{algorithm}[H]
\caption{Training Algorithm for MGCN: Train Masked Graph Convolutional Network}
\label{alg:mgcn_training}
\scriptsize
\begin{algorithmic}[0]
\Require Observed speeds $\Omega_{\text{obs}} \in \mathbb{R}^{n \times T}$, DFT speeds $\Omega_{\text{dft}} \in \mathbb{R}^{n \times T}$, mask $\mathcal{M} \in \{0,1\}^{n \times T}$, adjacency $A$, layers $L$, $\lambda_p$
\Ensure Trained parameters: $W_n^{(\ell)}, W_s^{(\ell)}$ for $\ell=1,\dots,L$, $W_{\text{out}}$, normalized $\hat{A}$
\State $\rightarrow$ Compute normalized adjacency $\hat{A} = \tilde{D}^{-1/2} \tilde{A} \tilde{D}^{-1/2}$ (add self-loops to $A$ if needed)
\State $\rightarrow$ Initialize weights $W_n^{(\ell)}, W_s^{(\ell)} \in \mathbb{R}^{d_{\ell-1} \times d_\ell}$, $W_{\text{out}} \in \mathbb{R}^{d_L \times 1}$
\For{each training epoch}
    \For{each time $t=1$ to $T-1$}
        \State $\rightarrow$ Form hybrid $\mathcal{H}^{(t)} = \mathcal{M}^{(t)} \odot \Omega_{\text{obs}}^{(t)} + (1 - \mathcal{M}^{(t)}) \odot \Omega_{\text{dft}}^{(t)}$
        \State $\rightarrow$ Set $\mathcal{H}^{(t,0)} = \mathcal{H}^{(t)}$
        \For{$\ell = 1$ to $L$}
            \State $\rightarrow$ Update $\mathcal{H}^{(t,\ell)} = \text{ReLU}\left( \hat{A} (\mathcal{M}^{(t)} \odot \mathcal{H}^{(t,\ell-1)}) W_n^{(\ell)} + \mathcal{H}^{(t,\ell-1)} W_s^{(\ell)} \right)$
        \EndFor
        \State $\rightarrow$ Predict $\hat{\Omega}^{(t+1)} = \mathcal{H}^{(t,L)} W_{\text{out}}$
        \State $\rightarrow$ Compute $\mathcal{L}_1 = \frac{1}{\sum \mathcal{M}^{(t+1)}} \sum \mathcal{M}^{(t+1)} (\hat{\Omega}^{(t+1)} - \Omega_{\text{obs}}^{(t+1)})^2$
        \State $\rightarrow$ Compute $\mathcal{L}_2 = \lambda_m |\hat{\Omega}^{(t+1)} - \Omega_{\text{dft}}^{(t+1)}|^2$
        \State $\rightarrow$ Total loss $\mathcal{L}_{gcn} = \mathcal{L}_1 + \mathcal{L}_2$
        \State $\rightarrow$ Update parameters via backpropagation
    \EndFor
\EndFor
\State $\rightarrow$ Return trained MGCN model
\end{algorithmic}
\end{algorithm}

\begin{algorithm}[H]
\caption{Computation Algorithm for MGCN: Real-Time Bus Arrival Time Estimation}
\label{alg:mgcn_computation}
\scriptsize
\begin{algorithmic}[0]
\Require Bus stop $(\vartheta_{\text{bs}}, \varphi_{\text{bs}})$, current bus $(\vartheta_{\text{bl}}, \varphi_{\text{bl}})$, graph $\mathcal{G}=(\mathcal{V},\mathcal{E})$, adjacency $A$, observed $\Omega_{\text{obs}}^{(t_s)}$, mask $\mathcal{M}^{(t_s)}$, trained MGCN ($W_n^{(\ell)}, W_s^{(\ell)}, W_{\text{out}}$), $\hat{A}$
\Ensure Estimated Time of Arrival ($\mathcal{ETA}_{MGCN}$)
\State $\rightarrow$ Identify $G_{\text{bl}} \in \mathcal{V}$ for $(\vartheta_{\text{bl}}, \varphi_{\text{bl}})$
\State $\rightarrow$ Identify $G_{\text{bs}} \in \mathcal{V}$ for $(\vartheta_{\text{bs}}, \varphi_{\text{bs}})$
\State $\rightarrow$ Compute MST-augmented graph from $\mathcal{G}$ to get $\mathcal{P}$
\State $\rightarrow$ Extract shortest path $\mathcal{P}' = (G_{\text{bl}} \rightsquigarrow G_{\text{bs}}) \subseteq \mathcal{P}$
\State $\rightarrow$ Initialize $\hat{\Omega}^{(t_s)} \gets \Omega_{\text{dft}}^{(t_s)}$ (or prior estimate)
\State $\rightarrow$ Compute hybrid $\mathcal{H}^{(t_s)} = \mathcal{M}^{(t_s)} \odot \Omega_{\text{obs}}^{(t_s)} + (1 - \mathcal{M}^{(t_s)}) \odot \hat{\Omega}^{(t_s)}$
\State $\rightarrow$ Set $\mathcal{H}^{(t_s,0)} = \mathcal{H}^{(t_s)}$
\For{$\ell = 1$ to $L$}
    \State $\rightarrow$ Update $\mathcal{H}^{(t_s,\ell)} = \text{ReLU}\left( \hat{A} (\mathcal{M}^{(t_s)} \odot \mathcal{H}^{(t_s,\ell-1)}) W_n^{(\ell)} + \mathcal{H}^{(t_s,\ell-1)} W_s^{(\ell)} \right)$
\EndFor
\State $\rightarrow$ Predict $\hat{\Omega}^{(t_s)} = \mathcal{H}^{(t_s,L)} W_{\text{out}}$
\State $\rightarrow$ Initialize $\mathcal{ETA}_{MGCN} \gets 0$
\For{each edge $(i,j) \in \mathcal{P}'$}
    \State $\rightarrow$ Compute $\mathcal{EL}_{ij} = \mathcal{H}\mathit{aversine}((\vartheta_i^{\text{median}}, \varphi_i^{\text{median}}), (\vartheta_j^{\text{median}}, \varphi_j^{\text{median}}))$
    \If{$\mathcal{M}^{(t_s)}_i = 1$ and $\mathcal{M}^{(t_s)}_j = 1$}
        \State $\rightarrow$ $\hat{\mathcal{U}}_{ij} \gets (\hat{\Omega}^{(t_s)}_i + \hat{\Omega}^{(t_s)}_j)/2$ \Comment{Use predicted if observed available}
    \Else
        \State $\rightarrow$ $\hat{\mathcal{U}}_{ij} \gets \hat{\Omega}^{(t_s)}_{ij}$ \Comment{For masked, use predicted}
    \EndIf
    \State $\rightarrow$ Update $\mathcal{ETA}_{MGCN} \gets \mathcal{ETA}_{MGCN} + \mathcal{EL}_{ij} / \hat{\mathcal{U}}_{ij}$
\EndFor
\State $\rightarrow$ Return $\mathcal{ETA}_{MGCN}$
\end{algorithmic}
\end{algorithm}

\begin{algorithm}[H]
\caption{Training Algorithm for HYB($n_m$): Train Hybrid Weighting Network}
\label{alg:hyb_nm_training}
\scriptsize
\begin{algorithmic}[0]
\Require Historical data: locations $(\vartheta_{t_s}, \varphi_{t_s})$, weather $\mathcal{W}_{t_s}$, times $\varsigma_{t_s}(m)$, speeds $v_{t_s}$, distances $\Delta_{t_s}^{bus-stop}$, actual ETAs $\mathcal{ETA}_{\text{actual}}$, model ETAs $\mathcal{ETA}_i$ ($i=1,\dots,5$)
\Ensure Trained neural network $f_{\text{NN}}$ for weights $\lambda'_{ih}$
\For{each timestamp $t_s$}
    \State $\rightarrow$ Compute sines: $\varsigma_{t_s}(m) = \sin(\alpha_{t_s} + 2m\pi / b)$ for $m=0,\dots,b-1$, $\alpha_{t_s} = 2\pi / 24 \cdot \text{hour}(t_s)$
    \State $\rightarrow$ Form feature $\mathbf{y}_{t_s} = (\vartheta_{t_s}, \varphi_{t_s}, \mathcal{W}_{t_s}, \varsigma_{t_s}(0), \dots, \varsigma_{t_s}(b-1), v_{t_s}, \Delta_{t_s}^{bus-stop})$
\EndFor
\State $\rightarrow$ Initialize $f_{\text{NN}}$ with final layer outputting $\chi_h(t_s) = (\chi_{1h}, \dots, \chi_{5h})$
\For{each training epoch}
    \For{each batch $\mathbf{y}_{t_s}$}
        \State $\rightarrow$ Predict raw weights $\chi_h(t_s) = f_{\text{NN}}(\mathbf{y}_{t_s})$
        \State $\rightarrow$ $\lambda_h(t_s) = Softmax(\chi_h{(t_s)})$ where,$ (\lambda_{ih}=e^{\chi_{ih}}/\sum_{j=1}^5e^{\chi_{jh}})$
        \State $\rightarrow$ Compute ETA: $\mathcal{ETA}_{hyb} = \sum_{i=1}^5 \lambda_{ih}.\mathcal{ETA}_i$
        \State $\rightarrow$ Loss $\mathcal{L}_{hyb} = (\mathcal{ETA}_{actual} - \mathcal{ETA}_{hyb})^2$
        \State $\rightarrow$ Update $f_{\text{NN}}$ parameters via backpropagation
    \EndFor
\EndFor
\State $\rightarrow$ Return trained $f_{\text{NN}}$
\end{algorithmic}
\end{algorithm}

\begin{algorithm}[H]
\caption{Computation Algorithm for HYB($n_m$): Hybrid ETA Prediction}
\label{alg:hyb_nm_computation}
\scriptsize
\begin{algorithmic}[0]
\Require Bus stop $(\vartheta_{\text{bs}}, \varphi_{\text{bs}})$, current bus $(\vartheta_{\text{bl}}, \varphi_{\text{bl}})$, weather $\mathcal{W}_{t_s}$, speed $v_{t_s}$, time $t_s$, trained models (MST-AV, GDRN-DFT, KNN, FE-NN, MGCN), trained $f_{\text{NN}}$, $n_m, b$
\Ensure Estimated Time of Arrival $\mathcal{ETA}'(n_m)_{\text{hyb}}$
\State $\rightarrow$ Compute distance $\Delta_{t_s}^{bus-stop} = from - (\vartheta_{\text{bl}}, \varphi_{\text{bl}})-to- (\vartheta_{\text{bs}}, \varphi_{\text{bs}})$ following $\mathcal{P'}$ from the network
\State $\rightarrow$ Compute sines: $\varsigma_{t_s}(m) = \sin(\alpha_{t_s} + 2m\pi / b)$ for $m=0,\dots,b-1$, $\alpha_{t_s} = 2\pi / 24 \cdot \text{hour}(t_s)$
\State $\rightarrow$ Form $\mathbf{y}_{t_s} = (\vartheta_{\text{bl}}, \varphi_{\text{bl}}, \mathcal{W}_{t_s}, \varsigma_{t_s}(0), \dots, \varsigma_{t_s}(b-1), v_{t_s}, \Delta_{t_s}^{bus-stop})$
\State $\rightarrow$ Compute raw weights $\lambda_h(t_s) = f_{\text{NN}}(\mathbf{y}_{t_s})$
\State $\rightarrow$ Identify top-$n_m$ indices in $\lambda_h(t_s)$, set mask $m_i = 1$ if $i$ in top-$n_m$, else $0$
\State $\rightarrow$ Normalize weights: $\lambda'_{ih}(t_s) = m_i \cdot \lambda_{ih}(t_s) / \sum_{j=1}^5 m_j \cdot \lambda_{jh}(t_s)$
\State $\rightarrow$ Compute model ETAs: $\mathcal{ETA}_{\text{MST-AV}}$, $\mathcal{ETA}_{\text{GDRN-DFT}}$, $\mathcal{ETA}_{\text{KNN}}$, $\mathcal{ETA}_{\text{FE-NN}}$, $\mathcal{ETA}_{\text{MGCN}}$
\State $\rightarrow$ Compute hybrid ETA: $\mathcal{ETA}'(n_m)_{\text{hyb}} = \sum_{i=1}^5 \lambda'_{ih} \cdot \mathcal{ETA}_i$
\State $\rightarrow$ Return $\mathcal{ETA}'(n_m)_{\text{hyb}}$
\end{algorithmic}
\end{algorithm}

\begin{algorithm}[H]
\caption{Computation Algorithm for Benchmark Spatio-Temporal Graph Models: Bus Arrival Time Estimation}
\label{alg:benchmark_eta_computation}
\scriptsize
\begin{algorithmic}[0]
\Require Test queries: bus locations $(\vartheta_{\text{bl}}, \varphi_{\text{bl}})$, stop locations $(\vartheta_{\text{bs}}, \varphi_{\text{bs}})$, current time $t$; imputed speeds $\Omega^{(t)} = \mathcal{M}^{(t)} \odot{\Omega}^{(t)}_{\text{obs}} + (1 - \mathbf{M}^{(t)}) \odot \mathbf{\Omega}^{(t)}_{\text{dft}} \in \mathbb{R}^{n \times T}$ (graph-level masked time series); graph $\mathcal{G}=(\mathcal{V}, \mathcal{E})$ with adjacency $\mathbf{A}$; trained benchmark models $m \in \{\text{DCRNN}, \text{STGCN}, \text{GWNet}, \text{T-GCN}, \text{MTGNN}, \text{ST-ResNet/DeepST}, \text{STFGNN},$\\$\text{ST-GConv/ChebNet}\}$
\Ensure Estimated Time of Arrival $\mathcal{ETA}_m$ for each benchmark model $m$
\State $\rightarrow$ Compute MST-augmented graph $\mathcal{P}$ from $\mathcal{G}$ (as in Alg.~\ref{alg:data_acq_preproc})
\For{each test query: $(\vartheta_{\text{bl}}, \varphi_{\text{bl}})$, $(\vartheta_{\text{bs}}, \varphi_{\text{bs}})$, $t$}
    \State $\rightarrow$ Identify grids $G_{\text{bl}}, G_{\text{bs}} \in \mathcal{V}$
    \State $\rightarrow$ Extract shortest path $\mathcal{P}' = (G_{\text{bl}} \rightsquigarrow G_{\text{bs}}) \subseteq \mathcal{P}$ using Haversine-weighted edges
    \For{each benchmark model $m$}
        \State $\rightarrow$ Forward pass on $\Omega_m^{(t)}$ to predict next-step speeds $\hat{\mathbf{\Omega}}^{(t+1)}_m \in \mathbb{R}^n$
        \Statex \hspace{1em} \textit{// Specific forward passes:}
        \Statex \hspace{1em} \textit{DCRNN:} Apply diffusion convolution (bidirectional kernels) + RNN on directed flows
        \Statex \hspace{1em} \textit{STGCN:} Spatial GCN + 1D temporal convolutions on $\mathbf{A}$
        \Statex \hspace{1em} \textit{GWNet:} Dilated temporal convolutions with fixed $\mathbf{A}$ (graph wavenet propagation)
        \Statex \hspace{1em} \textit{T-GCN:} GCN spatial embedding + GRU temporal recurrence
        \Statex \hspace{1em} \textit{MTGNN:} Mix-hop attention + dilated convolutions with fixed $\mathbf{A}$
        \Statex \hspace{1em} \textit{ST-ResNet/DeepST:} Grid-based residual convolutions
        \Statex \hspace{1em} \textit{STFGNN:} Fused spatial (GCN) and temporal (conv/GRU) layers
        \Statex \hspace{1em} \textit{ST-GConv/ChebNet:} Spectral GCN via Chebyshev polynomials on normalized Laplacian
        \State $\rightarrow$ Initialize $\mathcal{ETA}_m \gets 0$
        \For{each edge $(i,j) \in \mathcal{P}'$}
            \State $\rightarrow$ Compute edge length $\mathcal{EL}_{ij} \gets \mathcal{H}\mathit{aversine}((\vartheta_i^{\text{median}}, \varphi_i^{\text{median}}), (\vartheta_j^{\text{median}}, \varphi_j^{\text{median}}))$
            \State $\rightarrow$ Compute average velocity $\widehat{\mathcal{U}}_{ij}(t+1) \gets (\hat{\Omega}^{(t+1)}_{m,i} + \hat{\Omega}^{(t+1)}_{m,j}) / 2$
            \State $\rightarrow$ Update $\mathcal{ETA}_m \gets \mathcal{ETA}_m + \mathcal{EL}_{ij} / \widehat{\mathcal{U}}_{ij}(t+1)$
        \EndFor
        \State $\rightarrow$ Store $\mathcal{ETA}_m$ for query
    \EndFor
\EndFor
\State $\rightarrow$ Return $\{\mathcal{ETA}_m \mid m \in \text{benchmarks}\}$ for all queries
\end{algorithmic}
\end{algorithm}
\clearpage

\section{Tables}\label{secA2}

Additional details on the performance analysis of Route 2 and Route 3 and the hyperparameter study for different models are given below in the tables.
\begin{table}[h!]
\centering
\caption{Route 2 : Performance analysis (MAE \%; MAPE \%, RMSE)}
\label{tab:performance_study_route2}
\fontsize{5.9}{5}\selectfont
\renewcommand{\arraystretch}{1.20}
\setlength{\tabcolsep}{1.8pt}
\begin{tabular}{@{}lccccccccccccccccc@{}}
\toprule
 & & \multicolumn{5}{c}{Normal Conditions} & \multicolumn{4}{c}{Bus Stop Groups} & \multicolumn{5}{c}{Extreme Weather} \\
\cmidrule(r){3-7} \cmidrule(lr){8-11} \cmidrule(l){12-16}
Model & Metric & 0-10 & 10-25 & 25-45 & 45+ & Overall & 1-2 & 3-4 & 5-6 & 7+ & 0-10 & 10-25 & 25-45 & 45+ & Overall \\
\midrule
MST-AV & MAE  & 5.0 & 7.0 & 10.2 & 13.4 & 9.8 & 5.6 & 8.8 & 11.4 & 14.1 & 7.4 & 9.4 & 10.2 & 14.6 & 11.1 \\
 & MAPE & 67.0 & 38.9 & 28.8 & 20.0 & 23.4 & 49.1 & 33.2 & 23.3 & 20.7 & 85.0 & 46.3 & 30.9 & 23.3 & 28.4 \\
 & RMSE & 6.6 & 8.6 & 12.5 & 16.9 & 11.9 & 7.3 & 11.2 & 14.0 & 17.6 & 9.0 & 11.5 & 12.1 & 17.1 & 13.1 \\
\midrule
GDRN-DFT & MAE  & 3.9 & 6.8 & 7.1 & 9.3 & 7.0 & 5.5 & 8.5 & 9.5 & 10.3 & 6.8 & 8.3 & 9.6 & 11.2 & 9.2 \\
 & MAPE & 45.9 & 34.4 & 19.2 & 13.1 & 15.9 & 43.0 & 30.8 & 19.2 & 15.6 & 71.4 & 36.5 & 25.1 & 16.7 & 22.9 \\
 & RMSE & 5.3 & 8.4 & 9.1 & 12.0 & 8.9 & 7.3 & 10.7 & 11.8 & 12.4 & 8.0 & 9.9 & 11.3 & 13.2 & 10.5 \\
\midrule
KOOP-NET & MAE  & 1.6 & 5.3 & 12.2 & 14.2 & 11.0 & 3.0 & 6.1 & 13.7 & 17.7 & 3.6 & 7.0 & 10.5 & 13.1 & 9.4 \\
 & MAPE & 21.0 & 32.2 & 34.0 & 22.1 & 26.7 & 23.3 & 24.7 & 29.2 & 26.6 & 45.7 & 33.7 & 26.8 & 20.9 & 22.6 \\
 & RMSE & 2.6 & 7.2 & 15.0 & 17.4 & 13.4 & 4.3 & 8.0 & 16.9 & 21.7 & 4.5 & 8.5 & 12.5 & 15.3 & 10.3 \\
\midrule
FE-NN & MAE  & 3.2 & 5.2 & 11.3 & 13.1 & 10.0 & 3.4 & 6.1 & 12.0 & 16.0 & 2.8 & 5.0 & 8.8 & 11.0 & 7.8 \\
 & MAPE & 40.1 & 31.0 & 30.1 & 19.5 & 23.4 & 27.3 & 23.0 & 24.2 & 22.7 & 28.0 & 29.1 & 23.6 & 17.3 & 21.9 \\
 & RMSE & 4.5 & 7.1 & 13.7 & 16.2 & 12.1 & 4.7 & 7.9 & 14.7 & 19.7 & 3.9 & 6.4 & 10.5 & 13.0 & 9.8 \\
\midrule
RT-GCN & MAE  & 3.2 & 4.3 & 12.0 & 17.2 & 8.5 & 3.0 & 5.9 & 13.2 & 17.8 & 3.9 & 5.8 & 8.9 & 14.3 & 9.0 \\
 & MAPE & 38.7 & 23.4 & 33.5 & 25.6 & 21.4 & 27.2 & 20.7 & 25.5 & 26.8 & 49.4 & 29.3 & 24.2 & 22.5 & 23.2 \\
 & RMSE & 4.3 & 5.8 & 14.5 & 21.3 & 10.8 & 4.2 & 7.6 & 16.9 & 21.8 & 5.1 & 7.3 & 10.7 & 16.9 & 11.0 \\
\midrule
HYB-COMB(5) & MAE  & 1.8 & 4.8 & 8.0 & 9.2 & 6.1 & 3.1 & 6.0 & 9.9 & 11.1 & 3.0 & 5.2 & 8.9 & 11.1 & 7.5 \\
 & MAPE & 26.9 & 26.9 & 21.2 & 14.3 & 16.1 & 24.1 & 21.2 & 20.0 & 16.6 & 31.5 & 29.2 & 23.7 & 16.8 & 20.4 \\
 & RMSE & 2.3 & 6.0 & 9.1 & 11.0 & 7.4 & 4.3 & 7.7 & 12.2 & 13.4 & 4.0 & 6.6 & 10.6 & 13.1 & 9.1 \\
\bottomrule
\end{tabular}
\end{table}

\begin{table}[h!]
\centering
\caption{Route 3 : Performance analysis (MAE \%; MAPE \%, RMSE)}
\label{tab:performance_study_route3}
\fontsize{5.4}{5}\selectfont
\renewcommand{\arraystretch}{1.90}
\setlength{\tabcolsep}{1.6pt}
\begin{tabular}{@{}lccccccccccccccccc@{}}
\toprule
 & & \multicolumn{5}{c}{Normal Conditions} & \multicolumn{4}{c}{Bus Stop Groups} & \multicolumn{5}{c}{Extreme Weather} \\
\cmidrule(r){3-7} \cmidrule(lr){8-11} \cmidrule(l){12-16}
Model & Metric & 0-10 & 10-25 & 25-45 & 45+ & Overall & 1-2 & 3-4 & 5-6 & 7+ & 0-10 & 10-25 & 25-45 & 45+ & Overall \\
\midrule
MST-AV & MAE  & 4.70 & 6.64 & 9.26 & 11.68 & 9.11 & 5.37 & 7.80 & 10.26 & 12.84 & 6.97 & 8.40 & 10.42 & 13.40 & 10.38 \\
 & MAPE & 68.2 & 40.2 & 28.9 & 19.7 & 24.0 & 51.2 & 33.2 & 23.5 & 21.1 & 88.2 & 46.8 & 33.6 & 24.0 & 29.6 \\
 & RMSE & 6.04 & 8.10 & 11.07 & 14.40 & 10.93 & 6.86 & 9.64 & 12.34 & 15.73 & 8.25 & 10.03 & 12.18 & 15.40 & 12.03 \\
\midrule
GDRN-DFT & MAE  & 3.81 & 6.30 & 6.89 & 8.16 & 6.72 & 4.69 & 7.92 & 8.41 & 9.95 & 6.27 & 8.16 & 9.25 & 10.34 & 9.44 \\
 & MAPE & 47.6 & 35.0 & 20.0 & 12.9 & 16.6 & 41.9 & 31.7 & 19.2 & 16.4 & 73.7 & 38.8 & 26.5 & 17.3 & 25.1 \\
 & RMSE & 5.04 & 7.65 & 8.70 & 10.31 & 8.43 & 6.06 & 9.76 & 10.25 & 11.80 & 7.31 & 9.56 & 10.66 & 11.98 & 10.62 \\
\midrule
KOOP-NET & MAE  & 1.34 & 5.06 & 10.39 & 13.38 & 10.02 & 2.63 & 5.97 & 12.21 & 15.16 & 3.60 & 6.76 & 9.13 & 11.99 & 10.15 \\
 & MAPE & 20.0 & 33.2 & 33.0 & 22.8 & 27.0 & 23.2 & 26.2 & 29.4 & 26.2 & 49.4 & 35.5 & 26.8 & 21.5 & 27.9 \\
 & RMSE & 2.12 & 6.75 & 12.42 & 16.21 & 11.96 & 3.66 & 7.69 & 14.74 & 18.17 & 4.53 & 8.04 & 10.65 & 13.75 & 11.21 \\
\midrule
FE-NN & MAE  & 3.14 & 4.61 & 9.71 & 12.02 & 9.45 & 2.86 & 5.74 & 11.75 & 14.23 & 2.56 & 5.01 & 7.73 & 10.70 & 8.91 \\
 & MAPE & 41.8 & 30.6 & 29.5 & 19.7 & 24.3 & 26.5 & 23.8 & 25.6 & 22.8 & 28.7 & 31.3 & 23.8 & 18.5 & 25.8 \\
 & RMSE & 4.37 & 6.07 & 11.52 & 14.51 & 11.27 & 3.89 & 7.35 & 14.17 & 17.12 & 3.49 & 6.29 & 9.02 & 12.46 & 11.56 \\
\midrule
RT-GCN & MAE  & 2.91 & 4.16 & 10.05 & 14.63 & 8.42 & 2.97 & 5.39 & 11.64 & 15.97 & 3.37 & 5.71 & 8.43 & 12.69 & 8.73 \\
 & MAPE & 39.1 & 24.3 & 32.2 & 25.0 & 22.6 & 28.8 & 21.1 & 25.5 & 27.1 & 48.8 & 31.3 & 25.3 & 22.8 & 24.8 \\
 & RMSE & 3.92 & 5.51 & 11.86 & 17.69 & 10.58 & 4.11 & 6.88 & 14.52 & 19.09 & 4.30 & 7.11 & 9.95 & 14.69 & 10.50 \\
\midrule
HYB-COMB(5) & MAE  & 1.70 & 4.60 & 7.20 & 8.60 & 5.40 & 2.68 & 5.46 & 8.78 & 10.53 & 2.72 & 5.15 & 7.87 & 10.41 & 8.07 \\
 & MAPE & 24.6 & 25.4 & 19.9 & 13.6 & 15 & 23.9 & 21.6 & 20.1 & 17.3 & 32.7 & 32.1 & 24.1 & 17.5 & 23.2 \\
 & RMSE & 2.10 & 5.70 & 8.50 & 10.00 & 6.70 & 3.71 & 6.97 & 10.67 & 12.59 & 3.65 & 6.45 & 9.21 & 12.08 & 9.75 \\
\bottomrule
\end{tabular}
\end{table}

\begin{table}[h!]
\centering
\caption{Hyperparameter study of MGCN on MAE (Min) and Computation Time (s) with varying history length $\Delta T_m$, sampling interval $\delta_m$, and number of layers $L_m$. The optimal point is highlighted.\\}
\label{tab:mgcn-ablation}
\fontsize{4.5}{4}\selectfont
\renewcommand{\arraystretch}{1.20}
\setlength{\tabcolsep}{1.3pt}
\begin{tabular}{ccccccccccccccccc}
\toprule
$\Delta T_m$ (days) & \multicolumn{4}{c}{$L_m=1$} & \multicolumn{4}{c}{$L_m=2$} & \multicolumn{4}{c}{$L_m=3$} & \multicolumn{4}{c}{$L_m=4$} \\
\cmidrule{2-5}\cmidrule{6-9}\cmidrule{10-13}\cmidrule{14-17}
& 5 min & 10 min & 15 min & 20 min & 5 min & 10 min & 15 min & 20 min & 5 min & 10 min & 15 min & 20 min & 5 min & 10 min & 15 min & 20 min \\
\midrule
1 & \makecell{11.5 \\ 0.271} & \makecell{11.83 \\ 0.9} & \makecell{12.234 \\ 1.8} & \makecell{12.65 \\ 2.808} & \makecell{10.769 \\ 0.305} & \makecell{10.96 \\ 0.908} & \makecell{11.41 \\ 1.78} & \makecell{11.75 \\ 2.77} & \makecell{10.332 \\ 0.295} & \makecell{10.58 \\ 1.009} & \makecell{10.978 \\ 1.93} & \makecell{11.39 \\ 2.937} & \makecell{10.665 \\ 0.406} & \makecell{10.863 \\ 1.059} & \makecell{11.34 \\ 1.98} & \makecell{11.75 \\ 2.99} \\
\midrule
5 & \makecell{9.195 \\ 0.271} & \makecell{8.893 \\ 0.94} & \makecell{9.473 \\ 1.78} & \makecell{9.949 \\ 2.76} & \makecell{8.633 \\ 0.33} & \makecell{8.257 \\ 0.88} & \makecell{8.93 \\ 1.792} & \makecell{9.276 \\ 2.839} & \makecell{8.144 \\ 0.33} & \makecell{7.787 \\ 0.988} & \makecell{8.53 \\ 1.943} & \makecell{8.89 \\ 2.91} & \makecell{8.308 \\ 0.37} & \makecell{8.02 \\ 1.101} & \makecell{8.74 \\ 1.963} & \makecell{9.1 \\ 2.97} \\
\midrule
7 & \makecell{8.2 \\ 0.292} & \makecell{8.98 \\ 0.85} & \makecell{9.188 \\ 1.76} & \makecell{9.558 \\ 2.824} & \makecell{7.86 \\ 0.29} & \makecell{8.13 \\ 0.91} & \makecell{8.632 \\ 1.81} & \makecell{8.991 \\ 2.83} & \makecell{7.83 \\ 0.28} & \makecell{\textbf{7.39} \\ 1.039} & \makecell{8.537 \\ 1.861} & \makecell{8.65 \\ 2.86} & \makecell{8.02 \\ 0.361} & \makecell{8.002 \\ 1.1} & \makecell{8.74 \\ 1.971} & \makecell{8.92 \\ 3.011} \\
\midrule
10 & \makecell{9.03 \\ 0.31} & \makecell{8.75 \\ 0.889} & \makecell{9.59 \\ 1.766} & \makecell{10.03 \\ 2.79} & \makecell{8.456 \\ 0.33} & \makecell{8.334 \\ 0.86} & \makecell{8.978 \\ 1.845} & \makecell{9.38 \\ 2.75} & \makecell{8.22 \\ 0.253} & \makecell{8.184 \\ 0.979} & \makecell{8.89 \\ 1.92} & \makecell{9.22 \\ 2.936} & \makecell{8.4 \\ 0.431} & \makecell{8.52 \\ 1.06} & \makecell{9.1 \\ 2.05} & \makecell{9.464 \\ 3.004} \\
\bottomrule
\end{tabular}
\end{table}

\begin{table}[h!]
\centering
\caption{Performance Metrics of MST-AV Across Different Grid Sizes.\\\\}
\label{tab:mst-av-performance}
\fontsize{7}{7}\selectfont
\renewcommand{\arraystretch}{1.00}
\setlength{\tabcolsep}{2.2pt}
\begin{tabular}{llcccccc}\\
\toprule
Metric & Unit(mtr x mtr) & 25x25 & 50x50 & 100x100 & 250x250 & 500x500 & 1000x1000 \\
\midrule
MAE & (minutes) & 8.3 & 8.0 & 8.4 & 10.8 & 15.0 & 18.8 \\
Avg. Computation Time & (ms) & 120.4 & 85.6 & 59.2 & 57.8 & 51.4 & 48.2 \\
\bottomrule
\end{tabular}
\end{table}
\clearpage
\end{appendices}

\clearpage

\bibliography{sn-bibliography}

\end{document}